%% file: main.tex
\PassOptionsToPackage{table,xcdraw,usenames,dvipsnames}{xcolor}
\documentclass[10pt]{fairmeta}
\usepackage{url}
\usepackage{amsmath,amssymb,amsthm,mathtools,mathrsfs,bbm}
\usepackage{booktabs,multirow,tabularx,makecell,threeparttable,colortbl,arydshln}
\usepackage{graphicx,float}
\usepackage{wrapfig}
\usepackage{subcaption}
\newcommand{\subfigcontext}[1]{{\fontsize{6.5}{7.2}\selectfont #1}}
\usepackage{enumitem}
\IfFileExists{algorithm2e.sty}{\usepackage[ruled,vlined]{algorithm2e}}{\newcommand{\SetKwInOut}[2]{}}
\usepackage{pifont}
\usepackage{listings}
\usepackage{xspace}

\newtheorem{lemma}{Lemma}
\newtheorem{proposition}{Proposition}

\newif\ifsofttabdata
\newcolumntype{N}{>{\ifsofttabdata\fontsize{8}{10}\selectfont\fi}c}

\DeclareCaptionFormat{softgrpotable}{\textit{#1.}\enspace#3}
\newcommand{\softtabtoprule}{\specialrule{0.9pt}{0pt}{0pt}}
\newcommand{\softtabheadrule}{\hline\hline}
\newcommand{\softtabbottomrule}{\specialrule{0.9pt}{0pt}{0pt}}

\newcommand{\obsbox}[1]{%
    \begin{tcolorbox}[colframe=metaaccent!75!black, colback=metabg, boxrule=0.5pt, arc=1mm, top=2pt, bottom=3pt, left=3pt, right=3pt,
  boxsep=1pt]
        #1
    \end{tcolorbox}
}

\definecolor{bittersweet}{rgb}{1.0, 0.44, 0.37}
\definecolor{mygreen}{rgb}{0.29, 0.7, 0.48}
\definecolor{grpomethodcolor}{HTML}{C85E10}
\definecolor{esmethodcolor}{HTML}{588D31}
\newcommand{\grpomethod}{\textbf{\textcolor{grpomethodcolor}{GRPO}}}
\newcommand{\esmethod}{\textbf{\textcolor{esmethodcolor}{ES}}}

\definecolor{demphcolor}{RGB}{144,144,144}

\definecolor{mygray}{gray}{0.4}
\definecolor{autopurple}{HTML}{7030A0}
\definecolor{dyna_yellow}{HTML}{BF9000}
\definecolor{adaptive_blue}{HTML}{0070C0}
\definecolor{darksalmon}{rgb}{0.91, 0.59, 0.48}
\definecolor{emerald}{rgb}{0.31, 0.78, 0.47}
\definecolor{green(pigment)}{rgb}{0.0, 0.65, 0.31}
\definecolor{amaranth}{rgb}{0.9, 0.17, 0.31}
\definecolor{iris}{rgb}{0.35, 0.31, 0.81}
\definecolor{uu}{rgb}{0.95, 0.51, 0.51}
\definecolor{spirodiscoball}{rgb}{0.06, 0.75, 0.99}
\definecolor{mygrey}{gray}{0.4}
\definecolor{QuestionColor}{rgb}{0.7, 0.1, 0.1}
\definecolor{AnswerColor}{rgb}{0.1, 0.5, 0.1}
\definecolor{ReasoningColor}{rgb}{0.1, 0.1, 0.7}
\definecolor{promptaccent}{HTML}{A65300}
\definecolor{promptbg}{HTML}{F5F5F5}
\definecolor{promptrole}{HTML}{009E73}
\definecolor{promptvariable}{HTML}{0072B2}

\newtcolorbox{promptbox}[1]{%
  enhanced,
  breakable,
  colframe=promptaccent,
  colback=promptbg,
  colbacktitle=promptaccent,
  coltitle=white,
  fonttitle=\bfseries,
  title=#1,
  boxrule=0.6pt,
  arc=2mm,
  outer arc=2mm,
  top=3pt,
  bottom=3pt,
  left=5pt,
  right=5pt,
  boxsep=1pt,
  before skip=5pt,
  after skip=5pt
}

\SetKwInOut{Input}{Input}\SetKwInOut{Output}{Output}

\hypersetup{
    colorlinks=true,
    linkcolor=metaaccent,
    citecolor=metaaccent,
    filecolor=magenta,
    urlcolor=metaaccent,
}

\title{Understanding Evolution Strategies for LLM Reasoning:\linebreak
Broader Reasoning Coverage than GRPO}

\author[1,*]{Yunpeng Ba}
\author[2,*]{Zhi Zheng}
\author[1]{Yue Xie}
\author[5]{Jiaqing Li}
\author[3]{Xialiang Tong}
\author[3]{Tao Zhong}
\author[3]{\protect\linebreak\mbox{Mingxuan Yuan}}
\author[4]{Zhichao Lu}
\author[1]{Xuyang Wu}
\author[1]{Zhenkun Wang}

\affiliation[1]{Southern University of Science and Technology}
\affiliation[2]{National University of Singapore}
\affiliation[3]{Huawei Noah's Ark Lab}
\affiliation[4]{City University of Hong Kong}
\affiliation[5]{Harbin Institute of Technology, Weihai}

\metadata[GitHub \protect\githubemoji]{\url{https://github.com/yunpengba7/understanding-es}}
\correspondence{\email{zhi.zheng@u.nus.edu}, \email{wangzhenkun90@gmail.com}}

\abstract{
Evolution Strategies (ES) have recently emerged as a memory-efficient post-training paradigm for LLM reasoning.
However, the optimization behavior of ES remains understudied, making it hard to define its advantage scope compared to mainstream post-training paradigms (e.g., Group Relative Policy Optimization (GRPO)).
By systematically investigating ES dynamics and mechanisms, this paper \textbf{first identifies a performance advantage of ES over GRPO}, theoretically and empirically showing that ES can lead to broader reasoning coverage, thereby better exploiting the reasoning capabilities of pretrained LLMs.
Theoretically, we show that verifier-projected Jensen--Shannon diversity across the ES population is helpful to higher Pass@K performances.
Empirically, unlike GRPO, which exhibits entropy collapse, ES improves Pass@1 while attaining higher Pass@K than GRPO. We further develop a sequential GRPO--ES training strategy that combines GRPO's strength in Pass@1 with ES's gains in Pass@K.
\textbf{Second,} we find that despite substantial whole-model parameter drift, the task-performance gains of ES are only contributed to a sparse subset of larger-magnitude updates. This functional sparsity suggests that large parameter movement need not imply widespread functional change, and held-out evaluations further show that it does not necessarily lead to catastrophic forgetting.
\textbf{Finally,} we study how hyperparameter design affects the effectiveness of ES, demonstrating that ES requires a smaller population size in a larger LLM.
These findings position ES as a distinct reasoning post-training paradigm rather than a less effective, memory-efficient alternative to GRPO.
}

\begin{document}

\maketitle

\begin{figure}[H]
    \centering
    \vspace{-5pt}
    \includegraphics[width=1\textwidth]{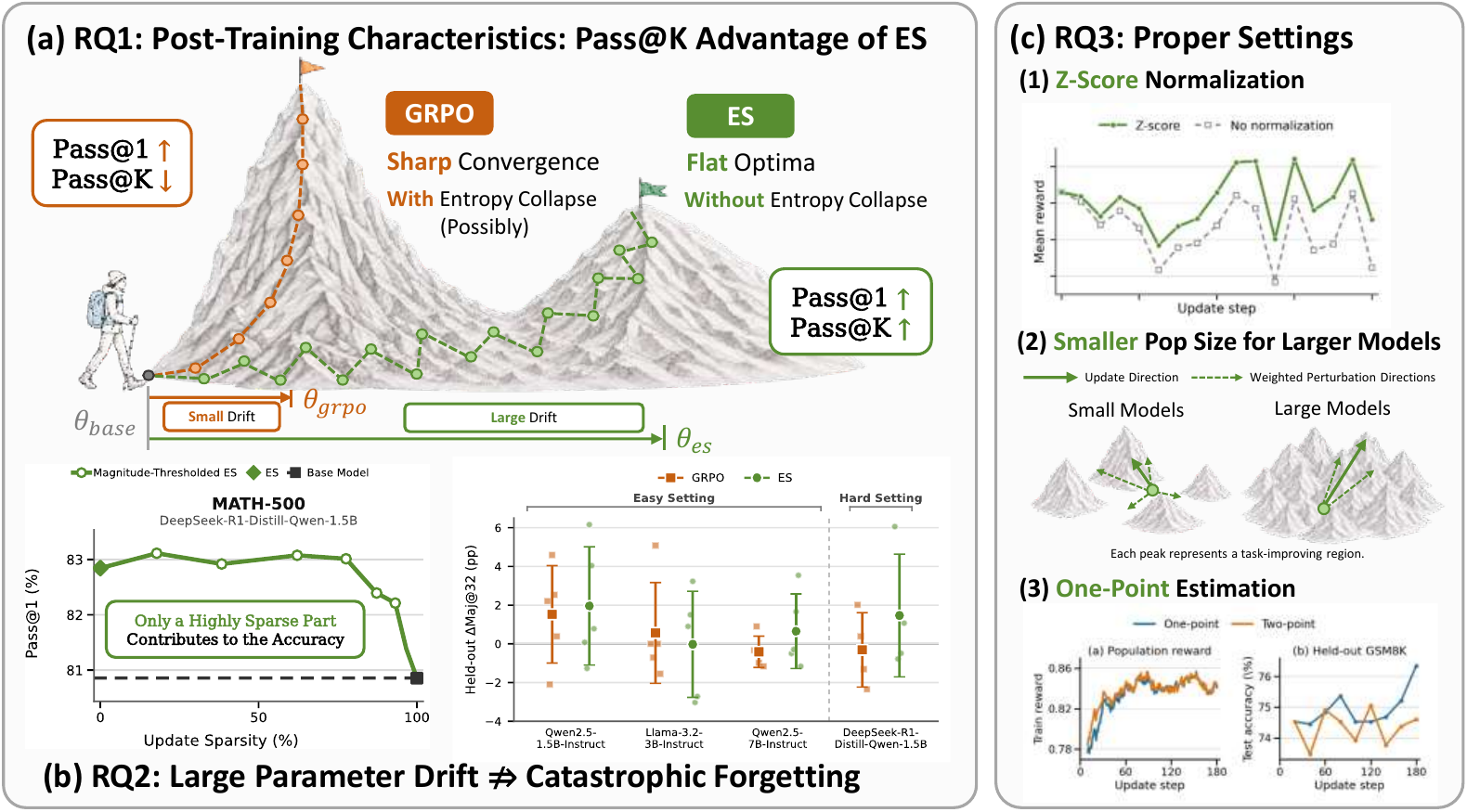}
    \caption{Overview of the three research questions and main findings. Panel (a) contrasts ES and GRPO post-training behavior; Panel (b) shows that keeping larger ES updates preserves target-task performance and reports held-out Maj@32 changes; Panel (c) summarizes normalization, population-size, and estimator choices.}
    \label{fig:rq123-overview}
    \vspace{-10pt}
\end{figure}

\section{Introduction}

Recent work has shown that Evolution Strategies (ES) \citep{salimans2017es} can fine-tune large language models (LLMs) to improve their reasoning capabilities \citep{qiu2025esatscale,sarkar2025hyperscale,zheng2026agentic}.
ES optimizes a model by perturbing its parameters, evaluating the perturbed models through forward passes, and aggregating reward-weighted perturbations to estimate an update direction.
By avoiding backpropagation, ES provides a memory-efficient and highly parallelizable approach to LLM post-training.
Despite these efficiency advantages, the practical viability of ES remains unclear: its post-training behavior, susceptibility to catastrophic forgetting \citep{hoy2026matching,abdi2026catastrophic}, and conditions for effective and scalable optimization are still poorly understood.

We compare ES with Group Relative Policy Optimization (GRPO) \citep{shao2024deepseekmath}, a mainstream RL method that likewise optimizes verifier rewards from sampled responses.
ES performs population-based search over parameter-perturbed policies, whereas GRPO samples multiple responses from a single policy and backpropagates a token-level objective based on their relative advantages.
While GRPO effectively improves Pass@1, it can exhibit entropy collapse, leading to increasingly concentrated exploration \citep{cui2025entropymechanism,petrenko2026entropy,jin2026revisiting}.
Consistently, prior analyses find that RL post-training can reduce reasoning-path diversity and even lower large-$K$ Pass@K relative to the base model \citep{yue2025doesrlincentivize,zhao2025echochamber,wu2025invisibleleash}.
This suggests that GRPO may concentrate probability on a narrower set of successful reasoning modes, making low-probability correct solutions less accessible.
We therefore investigate whether ES exhibits the same narrowing behavior or preserves broader reasoning coverage through population-based parameter-space exploration.

Beyond reasoning coverage, another important criterion for post-training is whether the model preserves its pretrained capabilities.
Compared with supervised fine-tuning (SFT), RL-based post-training methods such as GRPO have been shown to better preserve pretrained capabilities and to be less prone to catastrophic forgetting \citep{jin2025rl,yuan2025mitigating}.
This makes capability preservation an important reference point when evaluating whether ES can serve as a practical alternative to mainstream RL post-training.

For ES, however, whether such capability preservation holds remains unclear.
Prior work observes substantial parameter drift alongside catastrophic forgetting under ES and attributes the latter to the former \citep{abdi2026catastrophic}.
Yet whether parameter drift itself causes forgetting remains unverified, as existing evidence is limited to specific tasks and small training sets.
Beyond capability preservation, deploying ES effectively also requires understanding how its hyperparameters and estimator designs affect optimization stability and scalability, particularly as model size increases.
Together with the distinct exploration behavior discussed above, these open questions motivate our systematic study of ES. So, this paper summarizes our three research questions and main findings in Figure~\ref{fig:rq123-overview} and as follows:

\begin{enumerate}[leftmargin=*]
    \item \textbf{RQ1: Does ES exhibit the same post-training characteristics as GRPO?}
    We find that ES maintains broader reasoning coverage than GRPO.
    Across models post-trained on GSM8K and DeepScaleR, ES improves Pass@1 while achieving higher Pass@\(K\) than GRPO, without exhibiting the same entropy collapse.
    Theoretically, we show that verifier-projected Jensen--Shannon diversity across the ES population improves repeated-sampling success and can translate into higher Pass@\(K\) in the ES-updated policy.
    We further develop two sequential compositions, ES$\rightarrow$GRPO and GRPO$\rightarrow$ES, that combine GRPO's strength in Pass@1 with ES's gains in Pass@\(K\).

    \item \textbf{RQ2: Does ES necessarily cause catastrophic forgetting?}
    By examining the distribution of parameter changes, we find that the task-relevant effects of ES are concentrated in a small subset of larger-magnitude updates, while most parameter changes contribute little after perturbation cancellation.
    This functional sparsity suggests that substantial whole-model drift need not correspond to widespread functional change. Consistently, held-out capabilities remain largely preserved under appropriate training settings, indicating that large parameter movement alone does not imply catastrophic forgetting and that prior observations are better explained by training-set overfitting.

    \item \textbf{RQ3: What hyperparameter settings and estimators make ES effective and scalable?}
    We systematically evaluate ES hyperparameters and estimator designs to identify stable and effective configurations. We find that z-score reward normalization is a key ingredient for effective ES training. Due to the discrete reward in reasoning, the two-point estimator commonly favored in zeroth-order SFT provides no advantage for ES. We further find that the population size required for effective optimization decreases as pretrained model scale increases.
\end{enumerate}

\section{Preliminaries}

Let $\theta$ denote the model parameters and $\mathcal{D}_{\mathrm{train}}$ the distribution of reasoning prompts.
Given $x\sim\mathcal{D}_{\mathrm{train}}$, an LLM typically generates a chain-of-thought (CoT) trajectory $c$ followed by a final answer $a$:
\begin{equation}
c\sim\pi_\theta(\cdot\mid x), \qquad
a\sim\pi_\theta(\cdot\mid x,c),
\end{equation}
and we denote the complete response as $y=(c,a)$.
A verifier assigns a scalar reward $r(x,y)$, yielding the objective as follows:
\begin{equation}
F(\theta)
=
\mathbb{E}_{x\sim\mathcal{D}_{\mathrm{train}}}
\mathbb{E}_{y\sim\pi_\theta(\cdot\mid x)}
[r(x,y)].
\label{eq:population-reward-objective}
\end{equation}
Several post-training paradigms have been used to improve LLM reasoning capabilities, including SFT, On-Policy Distillation (OPD), GRPO, and ES. These methods differ primarily in the supervision used to optimize the generated reasoning trajectories.
\subsection{SFT \& OPD Fine-tuning}
SFT learns from expert CoT trajectories \citep{chu2025sft,zheng2025reasoning}, while OPD learns from token distributions of stronger teacher models \citep{song2026survey}.
Both require supervision beyond scalar verifier rewards, so we exclude them from our main comparison.

\subsection{GRPO Fine-tuning}

GRPO instead learns directly from verifier rewards.
We use GRPO as the primary gradient-based comparator to ES, as both optimize Eq.~\eqref{eq:population-reward-objective} but use fundamentally different update mechanisms.

For each reasoning prompt $x$, GRPO \citep{shao2024deepseekmath} samples a group of $G$ responses
$\{y_i=(c_i,a_i)\}_{i=1}^{G}$ from the behavior policy
$\pi_{\theta_{\mathrm{old}}}$ and obtains verifier rewards
$r_i=r(x,y_i)$.
Let $\bar r_G$ and $s_G$ denote the mean and population standard deviation of the group rewards.
For $s_G>0$, GRPO computes the response-level advantage
\begin{equation}
\widehat{A}_i
=
\frac{r_i-\bar r_G}{s_G}.
\label{eq:grpo-group-advantage}
\end{equation}
It then applies the same $\widehat{A}_i$ to all tokens in $y_i$ through a PPO-style clipped objective \citep{schulman2017ppo}.
Let $T_i=|y_i|$ and
\[
\varrho_{i,t}(\theta)
=
\frac{
\pi_{\theta}(y_{i,t}\mid x,y_{i,<t})
}{
\pi_{\theta_{\mathrm{old}}}(y_{i,t}\mid x,y_{i,<t})
}.
\]
The GRPO objective is
\begin{equation}
\mathcal{L}_{\mathrm{GRPO}}(\theta)
=
-\frac{1}{G}
\sum_{i=1}^{G}
\frac{1}{T_i}
\sum_{t=1}^{T_i}
\min\!\left(
\varrho_{i,t}\widehat{A}_i,\,
\operatorname{clip}
(\varrho_{i,t},1-\varepsilon_{\mathrm{clip}},1+\varepsilon_{\mathrm{clip}})
\widehat{A}_i
\right).
\label{eq:grpo-clipped-objective}
\end{equation}
The group reward provides a prompt-specific baseline without a learned critic, while each response-level advantage supervises the entire CoT trajectory and final answer.
\paragraph{Entropy Collapse.}
A known issue of GRPO-style policy-gradient training is \emph{entropy collapse}, where the policy distribution becomes increasingly concentrated during optimization.
For a tabular softmax policy at state $s$, let $p_a=\pi_\theta(a\mid s)$ and $A_a=A(s,a)$, with
$\mathbb{E}_{a\sim\pi_\theta}[A_a]=0$.
For policy entropy
$H(s)=-\sum_a p_a\log p_a$, prior work \citep{cui2025entropymechanism} gives the local relation
\begin{equation}
\Delta H(s)
\approx
-\eta_{\mathrm{PG}}\,
\operatorname{Cov}\!\left(\log p_a,p_aA_a\right).
\label{eq:grpo-entropy-dynamics}
\end{equation}
When high-probability actions tend to receive positive advantages, the covariance becomes positive, yielding $\Delta H(s)<0$ \citep{cui2025entropymechanism,petrenko2026entropy}.
Repeated updates can therefore reduce policy entropy and response diversity \citep{jin2026revisiting}.

\paragraph{Correct Answers under Repeated Sampling.}
Entropy measures overall distributional concentration, but does not directly show whether repeated sampling can produce a correct response.
We therefore use Pass@K, which measures whether at least one of $K$ sampled responses is correct.
Recent work shows that GRPO can reduce Pass@K relative to the Base Model, indicating a lower probability of finding a correct response through repeated sampling even when single-sample accuracy improves \citep{yue2025doesrlincentivize}.

\subsection{ES Optimization}

ES optimizes $F(\theta)$ through parameter-space evaluations \citep{salimans2017es}.
For perturbation scale $\sigma>0$, it considers the Gaussian-smoothed objective as follows:
\[
F_{\sigma}(\theta)
=
\mathbb{E}_{\epsilon\sim\mathcal{N}(0,I)}
[F(\theta+\sigma\epsilon)].
\]
At each update, ES samples $N$ perturbations $\{\epsilon_i\}_{i=1}^{N}$ and evaluates the corresponding perturbed models $\theta+\sigma\epsilon_i$, obtaining rollout rewards $R_i$. The standard one-point estimator is as follows:
\begin{equation}
\widehat{g}_{\mathrm{ES}}^{\mathrm{raw}}
=
\frac{1}{N\sigma}\sum_{i=1}^{N}R_i\epsilon_i,
\qquad
\mathbb{E}\!\left[\widehat{g}_{\mathrm{ES}}^{\mathrm{raw}}\right]
=
\nabla F_{\sigma}(\theta).
\label{eq:es-raw-estimator}
\end{equation}
In practice, following \citet{qiu2025esatscale}, we standardize rewards within each population, analogous to the group normalization in Eq.~\eqref{eq:grpo-group-advantage}.
Let $z_i$ denote the standardized reward.
The ES search direction and center-model update are as follows:
\begin{equation}
\widehat{d}_{\mathrm{ES}}
=
\frac{1}{N}\sum_{i=1}^{N}z_i\epsilon_i,
\qquad
\theta^+
=
\theta+\alpha\widehat{d}_{\mathrm{ES}},
\label{eq:es-practical-update}
\end{equation}
where $\alpha>0$ is the update scale and absorbs the fixed $1/\sigma$ factor.

\begin{wraptable}{r}{0.62\columnwidth}
    \vspace{-0.75\baselineskip}
    \centering
    {\fontsize{8}{9.5}\selectfont
    \setlength{\tabcolsep}{4pt}
    \begin{tabularx}{\linewidth}{@{}l>{\raggedright\arraybackslash}X>{\raggedright\arraybackslash}X@{}}
        \toprule[0.5mm]
        Aspect & ES & GRPO \\
        \midrule
        Signal path
        & $R_i\!\rightarrow\!z_i\!\rightarrow\!\theta$
        & $r_i\!\rightarrow\!\widehat A_i\!\rightarrow\!\theta$ \\
        Update rule
        & Standardized perturbation average
        & Clipped policy-gradient surrogate \\
        Execution
        & Perturbed-policy rollouts; no backprop
        & Policy rollouts and backprop \\
        Backward state
        & Not retained
        & Retained on GPU \\
        \bottomrule[0.5mm]
    \end{tabularx}
    \par}
    \caption{Update and efficiency comparison of ES and GRPO.}\vspace{-5pt}
    \label{tab:es-grpo-update-comparison}
\end{wraptable}

\paragraph{GRPO versus ES.} Table~\ref{tab:es-grpo-update-comparison} highlights the key distinction: GRPO backpropagates token-level advantages through a single policy, whereas ES converts population-level reward differences directly into a parameter-space update.

This mechanism-level distinction motivates our comparison of post-training behavior, while broader concerns about capability preservation and practical optimization motivate two additional questions. Accordingly, we investigate three questions about ES: \textbf{(1)} Does ES exhibit the same post-training characteristics as GRPO, particularly in terms of reasoning coverage? \textbf{(2)} Does large parameter movement under ES necessarily lead to catastrophic forgetting? \textbf{(3)} What hyperparameter settings and estimators make ES effective and scalable? The following sections investigate these questions in turn.

\section{RQ1: Does ES Exhibit the Same Post-Training Characteristics as GRPO?}

RQ1 examines whether ES can improve Pass@1 without degrading Pass@\(K\) at larger sampling budgets, unlike the entropy collapse and large-\(K\) Pass@\(K\) degradation that can occur under GRPO.

\subsection{Theoretically, How ES Population Diversity Can Support Pass@$K$}
\label{sec:rq1-theoretical-results}

Theoretically, we find that population diversity in ES can increase the probability that at least one of multiple sampled responses is correct, and that this advantage can transfer to the updated center policy under suitable conditions. The reason is that parameter perturbations induce heterogeneous policies with different success probabilities; sampling across these policies increases the chance of discovering a correct response, while reward weighting can preferentially emphasize more successful members. This subsection formalizes these effects and characterizes the conditions under which the resulting population-level advantage is preserved after the ES update.

\paragraph{ES Perturbations Induce Policy Diversity.}
We first quantify how ES parameter perturbations translate into diversity at the policy level.
To characterize the diversity induced by parameter perturbations, let
$s_\theta(y\mid x)=\nabla_\theta\log\pi_\theta(y\mid x)$
and $\pi_\theta^x=\pi_\theta(\cdot\mid x)$.
For a local displacement $\delta$, the prompt-conditioned Fisher information is
\begin{gather}
\mathcal{I}_x(\theta)
=
\mathbb{E}_{Y\sim\pi_\theta(\cdot\mid x)}
\left[s_\theta(Y\mid x)s_\theta(Y\mid x)^\top\right],
\notag\\
D_{\mathrm{KL}}
\left(\pi_{\theta+\delta}^x\,\|\,\pi_\theta^x\right)
=
\frac{1}{2}\delta^\top\mathcal{I}_x(\theta)\delta
+o(\|\delta\|^2).
\label{eq:rq1-fisher-geometry}
\end{gather}
For $\delta=\sigma\epsilon$, the expected local policy displacement is proportional to
$\frac{\sigma^2}{2}\operatorname{tr}\mathcal{I}_x(\theta)$.

Let $\mathcal{D}$ be the evaluation-prompt distribution, let $v(x,y)\in\{0,1\}$ be a correctness verifier, and define the correct-response set $\mathcal{C}(x)=\{y:v(x,y)=1\}$.
For any policy $\pi$, let $p_\pi(x)=\Pr_{Y\sim\pi(\cdot\mid x)}[Y\in\mathcal{C}(x)]$.
Unless an expectation is shown, quantities below are conditioned on the realized population, fitness values, and resulting update. So, we provide the lemma as follows:

\begin{lemma}[Perturbations induce policy diversity]
\label{lem:rq1-local-policy-js}
For a fixed prompt $x$ and population size $N$, draw
$\epsilon_i\overset{\mathrm{iid}}{\sim}\mathcal{N}(0,I)$, define
$\pi_i(y\mid x)=\pi_{\theta+\sigma\epsilon_i}(y\mid x)$ and
$\bar\pi_N=N^{-1}\sum_i\pi_i$, and let
$\operatorname{JS}^{\mathrm{pol}}_N(x)
=N^{-1}\sum_iD_{\mathrm{KL}}(\pi_i\,\|\,\bar\pi_N)$.
The prompt-conditioned Fisher information $\mathcal{I}_x(\theta)$ is defined in Equation~\eqref{eq:rq1-fisher-geometry}.
Assuming finite entropy and Fisher information, common local support, and sufficient smoothness as detailed in Appendix~\ref{app:proof-local-policy-js}, as $\sigma\to0$,
\begin{equation}
\mathbb{E}_{\epsilon_{1:N}}\!\left[\operatorname{JS}^{\mathrm{pol}}_N(x)\right]
=\frac{\sigma^2}{2}\!\left(1-\frac{1}{N}\right)
\operatorname{tr}\mathcal{I}_x(\theta)+O(\sigma^4).
\label{eq:rq1-local-policy-js}
\end{equation}
\end{lemma}

\paragraph{Multiple Policies in the ES Population Increase the Chance of Finding a Correct Answer.}
The benefit of sampling across different ES members is then established through comparison with matched sampling from a single policy.
\begin{lemma}[Policy diversity improves correct-answer discovery]
\label{lem:rq1-population-coverage}
For these policies, define
$p_i(x)=p_{\pi_i}(x)$
and $\bar p(x)=N^{-1}\sum_i p_i(x)$.
Using binary entropy $h$, define
$\operatorname{JS}^{\mathrm{succ}}_N(x)
=h(\bar p(x))-N^{-1}\sum_i h(p_i(x))$.
Data processing, followed by one independent sample per member, yields
\begin{gather}
0
\le
\operatorname{JS}^{\mathrm{succ}}_N(x)
\le
\operatorname{JS}^{\mathrm{pol}}_N(x),
\label{eq:rq1-js-data-processing}\\[-0.25ex]
P_N^{\mathrm{pop}}(x)
:=1-\prod_{i=1}^N(1-p_i(x))
\ge 1-(1-\bar p(x))^N=:P_N^{\mathrm{same}}(x).
\label{eq:rq1-population-coverage-bound}
\end{gather}
Here $P_N^{\mathrm{pop}}$ and $P_N^{\mathrm{same}}$ denote one-response-per-member success and its matched single-policy baseline.
Equality holds if and only if $p_1(x)=\cdots=p_N(x)$, and the local gap is proportional to $\operatorname{JS}^{\mathrm{succ}}_N(x)$ as derived in Appendix~\ref{app:proof-population-coverage}.
\end{lemma}

\paragraph{Reward Weighting Can Exploit Population Heterogeneity.}
The condition under which reward weighting improves the population-level success rate is then characterized.
\begin{lemma}[Reward weighting improves success under positive correlation]
\label{lem:rq1-selection-gain}
Let $w_i\ge0$, $\sum_iw_i=1$, be normalized weights obtained by a monotone transformation of realized fitness $R_i$.
Define the reward-weighted policy mixture $\pi_w=\sum_iw_i\pi_i$ and its success probability $p_w(x)=\sum_iw_ip_i(x)$.
With $\operatorname{Cov}_i$ taken under a uniform member draw,
$p_w(x)-\bar p(x)=N\operatorname{Cov}_{i}(w_i,p_i(x))$.
\end{lemma}

The analytical comparator $\pi_w$ improves over the uniform population average exactly when the weights and member success are positively correlated.
It is not the practical ES update or a comparison against $\pi_\theta$.

\paragraph{Sufficient Conditions for Pass@$K$ Improvement of the ES Center.}
Finally, sufficient conditions are established under which the ES-updated model achieves higher Pass@\(K\) than the model before the update.
\begin{proposition}[ES updates can improve Pass@\(K\)]
\label{prop:rq1-center-transfer}
Let $B_w(x)$ and $B_{\theta^+}(x)$ denote Bernoulli outcome distributions with success probabilities $p_w(x)$ and $p_{\theta^+}(x)$, respectively.
For a repeated-sampling budget $K$, let
$J_K(\pi)=\mathbb{E}_{x\sim\mathcal{D}}[1-(1-p_\pi(x))^K]$ denote expected Pass@\(K\).
Define the comparator margin $m_K=J_K(\pi_w)-J_K(\pi_\theta)$, and let $\varepsilon_{\mathrm{succ}}\ge0$ be the center-transfer error budget.
Assume
\begin{gather}
\mathbb{E}_{x\sim\mathcal{D}}D_{\mathrm{KL}}
\!\left(B_w(x)\,\|\,B_{\theta^+}(x)\right)
\le\varepsilon_{\mathrm{succ}},
\label{eq:rq1-binary-transfer}\\[-0.25ex]
m_K>K\sqrt{\varepsilon_{\mathrm{succ}}/2}.
\label{eq:rq1-positive-transfer-margin}
\end{gather}
Then
\begin{equation}
J_K(\pi_{\theta^+})
\ge J_K(\pi_w)-K\sqrt{\varepsilon_{\mathrm{succ}}/2}
>J_K(\pi_\theta).
\label{eq:rq1-center-transfer-bound}
\end{equation}
\end{proposition}

Thus, when verifier-relevant heterogeneity improves population coverage, reward-aligned selection produces a sufficient margin over the initial policy, and the center update preserves this margin; ES training can improve the model's Pass@\(K\).
Proofs and implementation-related qualifications are provided in Appendix~\ref{app:proof-local-policy-js}--\ref{app:proof-center-transfer}.

\subsection{Empirically, ES Achieves Higher Pass@\(K\) Than GRPO and Base LLM}
\label{sec:rq1-empirical-comparison}

\textbf{Experimental Setup.}
In this part, we empirically evaluate whether ES can improve Pass@1 while preserving Pass@\(K\) under two post-training settings.
The Easy Setting compares ES and GRPO on Qwen2.5-1.5B-Instruct, Llama-3.2-3B-Instruct, and Qwen2.5-7B-Instruct after two epochs of GSM8K post-training \citep{cobbe2021gsm8k}.
The Hard Setting compares them on DeepSeek-R1-Distill-Qwen-1.5B after one epoch of DeepScaleR post-training \citep{luo2025deepscaler}.
We FLOP-match ES and GRPO by using \(N=32\) perturbation directions for ES and \(G=8\) responses per prompt for GRPO. Appendix~\ref{app:es-grpo-flop-matching} gives the complete accounting.

\begin{figure}[t]
    \centering
        \vspace{-10pt}
    \begin{minipage}[t]{0.5\linewidth}
        \centering
        \vspace*{-3pt}
        {\scriptsize\textbf{Training Process:} GSM8K $\rightarrow$ GPQA\par}
        \vspace{1pt}
        \includegraphics[height=0.20in]{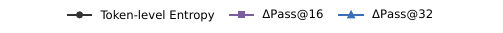}\par
    \end{minipage}%
    \begin{minipage}[t]{0.5\linewidth}
        \centering
        \vspace*{-3pt}
        {\scriptsize\textbf{Test Results:} Training Task $\rightarrow$ Test Task\par}
        \vspace{1pt}
        \includegraphics[height=0.20in]{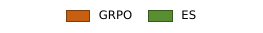}\par
    \end{minipage}
    \vspace{-4pt}

    \begin{subfigure}[t]{0.25\linewidth}
        \centering
        \includegraphics[width=\linewidth]{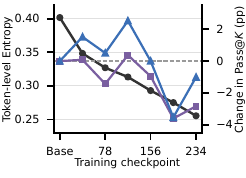}\par
        \caption{\textbf{GRPO}\\[1pt]
        \subfigcontext{Qwen2.5-1.5B-Instruct}}
        \label{fig:rq1-overview-grpo}
    \end{subfigure}%
    \begin{subfigure}[t]{0.25\linewidth}
        \centering
        \includegraphics[width=\linewidth]{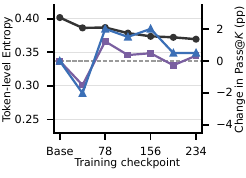}\par
        \caption{\textbf{ES}\\[1pt]
        \subfigcontext{Qwen2.5-1.5B-Instruct}}
        \label{fig:rq1-overview-es}
    \end{subfigure}%
    \begin{subfigure}[t]{0.25\linewidth}
        \centering
        \includegraphics[width=\linewidth]{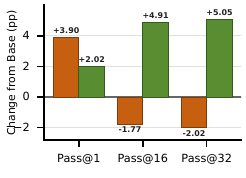}\par
        \caption{\textbf{GSM8K \(\rightarrow\) GPQA}\\[1pt]
        \subfigcontext{Llama-3.2-3B-Instruct}}
        \label{fig:rq1-overview-easy-gpqa}
    \end{subfigure}%
    \begin{subfigure}[t]{0.25\linewidth}
        \centering
        \includegraphics[width=\linewidth]{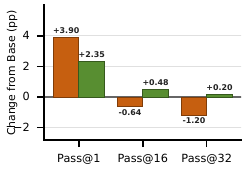}\par
        \caption{\textbf{DeepScaleR \(\rightarrow\) MATH-500}\\[1pt]
        \subfigcontext{DeepSeek-R1-Distill-Qwen-1.5B}}
        \label{fig:rq1-overview-hard-math500}
    \end{subfigure}
    \vspace{-3pt}
    \caption{
    Comparison of ES and GRPO during training and testing.
    During GSM8K post-training, GPQA token-level entropy drops substantially under GRPO but remains largely stable under ES; GRPO finishes below the base model on Pass@16 and Pass@32, whereas ES finishes above it.
    On the representative task pairs, GRPO raises Pass@1 but lowers Pass@16 and Pass@32, whereas ES improves all three metrics.
    }
        \vspace{-10pt}
    \label{fig:rq1-overview}
\end{figure}

\textbf{Training Dynamics.}
Figure~\ref{fig:rq1-overview}(a--b) tracks held-out GPQA performance during GSM8K post-training of Qwen2.5-1.5B-Instruct.
Under GRPO, token-level entropy declines substantially, while Pass@16 and Pass@32 finish below the base model.
This pattern is consistent with prior reports linking entropy collapse under GRPO-style optimization to reduced reasoning diversity and weaker Pass@\(K\) scaling \citep{jang2026badapples}.
In contrast, under ES, entropy changes modestly, and both metrics finish above the base model.

\textbf{Test Results.}
The representative task pairs in Figure~\ref{fig:rq1-overview}(c--d) expose the Pass@1--Pass@\(K\) difference directly.
On GPQA after GSM8K post-training and MATH-500 after DeepScaleR post-training, GRPO improves Pass@1 but falls below the corresponding base model on Pass@16 and Pass@32, whereas ES improves all three metrics.
The complete results in Tables~\ref{tab:rq1-instruct-final} and~\ref{tab:rq1-deepseek-math} show that ES improves average Pass@1, Pass@16, and Pass@32 relative to the base model in both settings.
Although GRPO achieves larger average Pass@1 gains, ES attains higher average Pass@16 and Pass@32 in both settings.
Pass@\(K\) degradation is particularly common in the Easy Setting: GRPO falls below the corresponding base model on both Pass@16 and Pass@32 in 15 of 18 comparisons.
These results indicate that ES yields gains at both \(K=1\) and larger \(K\), whereas GRPO's gains are concentrated at \(K=1\) and often accompanied by lower large-\(K\) performance.
Concurrent work reports a similar observation: ES achieves higher Pass@\(K\) than RL and preserves solution coverage, whereas RL plateaus earlier and can be overtaken by the Base Model as \(K\) increases \citep{hayes2026beyond}.
% Their averages increase in the Hard Setting despite declining on the representative MATH-500 task, showing that aggregate results can mask task-level variation.
% The average modal accuracies of ES and GRPO remain close: their largest difference is \(1.41\) percentage points, observed at Maj@16 in the Hard Setting.
% Higher Pass@\(K\) with comparable Maj@\(K\) is consistent with ES retaining less frequently sampled correct reasoning paths without degrading the modal answer.

\textbf{Methodology: Sequentially Mixed Training of ES and GRPO}
ES leads in Pass@$K$ over base and GRPO, while GRPO improve much Pass@1 than ES. So, to combine their respective advantages, we propose an intuitive method by sequentially mixing the training process of ES and GRPO. Under the same total update budget, we additionally split training equally between two stages and evaluate both sequential compositions: ES$\rightarrow$GRPO applies ES first and GRPO second, whereas GRPO$\rightarrow$ES reverses the order.
Figure~\ref{fig:rq1-overview} presents a representative training trajectory and two representative training-to-test task pairs, while Tables~\ref{tab:rq1-instruct-final} and~\ref{tab:rq1-deepseek-math} report the complete Pass@\(K\) results.
The corresponding Maj@\(K\) results are reported in Appendix~\ref{app:majority-vote-results}.
For each prompt, Pass@1 measures single-sample correctness, Pass@\(K\) records whether at least one of \(K\) independent responses is correct, and Maj@\(K\) reports majority-vote accuracy after answer normalization.

\begin{table*}[t]
\centering
{\fontsize{8}{10}\selectfont

% Table height

% =========================================================
% Pass Metrics
% =========================================================
\setlength{\tabcolsep}{3pt}

\global\softtabdatafalse
\resizebox{\textwidth}{!}{%
\begin{tabular}{@{}l*{21}{N}@{}}
\multicolumn{22}{c}{\textbf{Pass Metrics}} \\
\toprule
& \multicolumn{3}{c}{GSM8K}
& \multicolumn{3}{c}{CSQA}
& \multicolumn{3}{c}{HotpotQA}
& \multicolumn{3}{c}{Countdown}
& \multicolumn{3}{c}{GPQA}
& \multicolumn{3}{c}{MBPP}
& \multicolumn{3}{c}{Average} \\
\cmidrule(lr){2-4}
\cmidrule(lr){5-7}
\cmidrule(lr){8-10}
\cmidrule(lr){11-13}
\cmidrule(lr){14-16}
\cmidrule(lr){17-19}
\cmidrule(lr){20-22}

Method
& @1 & @16 & @32
& @1 & @16 & @32
& @1 & @16 & @32
& @1 & @16 & @32
& @1 & @16 & @32
& @1 & @16 & @32
& @1 & @16 & @32 \\
\midrule

\multicolumn{22}{l}{\textbf{Qwen2.5-1.5B-Instruct}} \\

\global\softtabdatatrue
Base
& 70.3 & 94.8 & 96.4
& 60.6 & 93.4 & 96.1
& \underline{27.6} & \underline{46.6} & \underline{50.3}
& 7.5 & 46.9 & 56.9
& 21.4 & 87.1 & 94.4
& 58.6 & \underline{83.8} & \underline{87.2}
& 41.0 & \cellcolor{gray!35}75.4 & \cellcolor{gray!35}80.2 \\

\grpomethod
& \underline{78.1} & 93.9 & 95.3
& \underline{63.7} & 93.2 & 95.5
& 25.0 & 44.3 & 48.1
& 8.5 & \underline{49.5} & 58.8
& 22.2 & 87.3 & 96.0
& \underline{60.0} & 82.4 & 86.0
& \textbf{42.9} & 75.1 & 79.9 \\

\esmethod
& 73.2 & \underline{94.9} & \underline{96.6}
& 60.7 & \underline{94.1} & \underline{96.8}
& 24.9 & 45.8 & 49.9
& \underline{8.8} & 49.4 & \underline{59.5}
& \underline{22.9} & \underline{88.5} & \underline{96.5}
& 58.7 & 83.0 & 86.4
& 41.5 & \textbf{76.0} & \textbf{80.9} \\

\esmethod$\rightarrow$\grpomethod
& 77.1 & 94.4 & 96.0
& 62.1 & 93.9 & 96.5
& 24.3 & 44.0 & 47.9
& 8.3 & 47.9 & 57.1
& 22.2 & 88.1 & 96.0
& 59.8 & 83.4 & 86.8
& \cellcolor{gray!35}42.3 & 75.3 & 80.0 \\

\grpomethod$\rightarrow$\esmethod
& 76.1 & 94.6 & 96.1
& 62.5 & 93.7 & 96.0
& 24.6 & 44.7 & 48.8
& 7.8 & 48.4 & 58.0
& 22.8 & 86.6 & 95.5
& 58.5 & 81.6 & 84.8
& 42.1 & 75.0 & 79.9 \\

\midrule
\multicolumn{22}{l}{\textbf{Llama-3.2-3B-Instruct}} \\

Base
& 77.7 & \underline{96.8} & \underline{98.1}
& 67.3 & 92.8 & 95.3
& 26.6 & 44.2 & 47.6
& 7.7 & 47.4 & 56.6
& 23.0 & 82.0 & 90.4
& \underline{62.1} & 80.6 & 83.7
& 44.1 & \cellcolor{gray!35}74.0 & \cellcolor{gray!35}78.6 \\

\grpomethod
& \underline{84.7} & 95.6 & 96.4
& \underline{72.1} & 90.5 & 92.8
& 27.5 & 40.4 & 43.0
& \underline{9.6} & \underline{52.9} & \underline{61.6}
& \underline{26.9} & 80.3 & 88.4
& 62.1 & 76.7 & 79.8
& \textbf{47.1} & 72.7 & 77.0 \\

\esmethod
& 81.6 & 96.4 & 97.9
& 69.9 & \underline{93.0} & \underline{95.5}
& \underline{29.0} & \underline{45.7} & \underline{48.9}
& 9.0 & 51.3 & 60.6
& 25.1 & \underline{86.9} & \underline{95.5}
& 61.1 & \underline{81.2} & \underline{84.1}
& 45.9 & \textbf{75.8} & \textbf{80.4} \\

\esmethod$\rightarrow$\grpomethod
& 84.1 & 95.0 & 96.2
& 72.0 & 91.3 & 93.7
& 26.1 & 41.0 & 44.1
& 10.3 & 53.5 & 63.4
& 24.4 & 71.4 & 79.3
& 63.5 & 80.4 & 82.1
& \cellcolor{gray!35}46.7 & 72.1 & 76.5 \\

\grpomethod$\rightarrow$\esmethod
& 84.1 & 95.9 & 96.9
& 71.3 & 91.8 & 94.3
& 27.5 & 43.6 & 47.0
& 8.5 & 49.9 & 59.5
& 26.6 & 84.5 & 92.9
& 61.5 & 77.2 & 80.5
& 46.6 & 73.8 & 78.5 \\

\midrule
\multicolumn{22}{l}{\textbf{Qwen2.5-7B-Instruct}} \\

Base
& 91.8 & 97.6 & 98.0
& 80.0 & \underline{92.9} & \underline{94.3}
& 45.4 & \underline{52.4} & \underline{53.6}
& 34.1 & 70.3 & 74.8
& 33.3 & 82.2 & \underline{89.4}
& \underline{76.1} & \underline{86.4} & \underline{87.9}
& 60.1 & \cellcolor{gray!35}80.3 & \cellcolor{gray!35}83.0 \\

\grpomethod
& \underline{92.3} & 97.0 & 97.4
& \underline{80.8} & 91.7 & 92.7
& \underline{46.2} & 52.3 & 53.3
& \underline{36.8} & 68.4 & 72.9
& \underline{34.8} & 79.3 & 85.4
& 74.8 & 85.5 & 87.6
& \textbf{61.0} & 79.0 & 81.5 \\

\esmethod
& 91.0 & \underline{97.6} & \underline{98.1}
& 79.0 & 92.6 & \underline{94.3}
& 45.1 & 52.1 & 53.4
& 33.0 & \underline{71.9} & \underline{76.7}
& 34.4 & \underline{83.6} & 88.9
& 75.1 & 85.7 & 87.6
& 59.6 & \textbf{80.6} & \textbf{83.1} \\

\esmethod$\rightarrow$\grpomethod
& 92.1 & 97.6 & 98.0
& 80.2 & 92.4 & 93.7
& 46.0 & 52.1 & 53.1
& 36.0 & 70.7 & 75.5
& 35.5 & 79.9 & 85.9
& 75.2 & 85.1 & 86.0
& \cellcolor{gray!35}60.8 & 79.7 & 82.0 \\

\grpomethod$\rightarrow$\esmethod
& 92.3 & 97.4 & 98.0
& 80.2 & 91.9 & 93.2
& 45.9 & 52.4 & 53.6
& 35.0 & 69.4 & 73.8
& 34.2 & 80.8 & 87.4
& 75.1 & 86.0 & \underline{87.9}
& 60.4 & 79.6 & 82.3 \\

\bottomrule
\end{tabular}%
}
\global\softtabdatafalse

\par}

\caption{
Pass@\(K\) results (\(\times 100\)) in the Easy Setting after two epochs of GSM8K post-training.
Underlining marks the best Base/GRPO/ES result in each task cell; bold and shading mark the best and second-best Average cells across all five methods.
}
\label{tab:rq1-instruct-final}

\end{table*}
\begin{table*}[t]
    \centering
    {\fontsize{8}{10}\selectfont

    % =========================================================
    % Pass Metrics
    % =========================================================
    \setlength{\tabcolsep}{8pt}
    \global\softtabdatafalse

    \resizebox{\textwidth}{!}{%
    \begin{tabular}{@{}l*{15}{N}@{}}
    \multicolumn{16}{c}{\textbf{Pass Metrics}} \\
    \toprule
    & \multicolumn{3}{c}{AIME24}
    & \multicolumn{3}{c}{AIME25}
    & \multicolumn{3}{c}{AMC23}
    & \multicolumn{3}{c}{MATH500}
    & \multicolumn{3}{c}{Average} \\
    \cmidrule(l{1pt}r{1pt}){2-4}
    \cmidrule(l{1pt}r{1pt}){5-7}
    \cmidrule(l{1pt}r{1pt}){8-10}
    \cmidrule(l{1pt}r{1pt}){11-13}
    \cmidrule(l{1pt}r{1pt}){14-16}

    Method
    & @1 & @16 & @32
    & @1 & @16 & @32
    & @1 & @16 & @32
    & @1 & @16 & @32
    & @1 & @16 & @32 \\
    \midrule

    \multicolumn{16}{l}{\textbf{DeepSeek-R1-Distill-Qwen-1.5B}} \\

    \global\softtabdatatrue
    Base
    & 23.4 & 59.9 & 66.7
    & 22.6 & 43.8 & 50.0
    & 64.1 & 93.6 & 95.0
    & 80.9 & 96.7 & 98.0
    & 47.7 & 73.5 & 77.4 \\

    \grpomethod
    & \underline{27.6} & \underline{62.2} & 66.7
    & \underline{25.4} & 45.8 & \underline{53.3}
    & \underline{73.9} & 94.7 & 95.0
    & \underline{84.8} & 96.1 & 96.8
    & \textbf{52.9} & 74.7 & 78.0 \\

    \esmethod
    & 22.8 & 60.3 & \underline{70.0}
    & 23.2 & \underline{46.7} & 50.0
    & 70.5 & \underline{96.0} & \underline{97.5}
    & 83.2 & \underline{97.2} & \underline{98.2}
    & 49.9 & 75.0 & \cellcolor{gray!35}78.9 \\

    \esmethod$\rightarrow$\grpomethod
    & 29.0 & 70.6 & 76.7
    & 23.8 & 43.0 & 50.0
    & 72.3 & 92.5 & 92.5
    & 84.4 & 97.1 & 97.8
    & \cellcolor{gray!35}52.3 & \cellcolor{gray!35}75.8 & \textbf{79.2} \\

    \grpomethod$\rightarrow$\esmethod
    & 25.8 & 61.8 & 63.3
    & 24.5 & 51.0 & 56.7
    & 71.6 & 94.9 & 95.0
    & 84.5 & 97.0 & 97.6
    & 51.6 & \textbf{76.2} & 78.2 \\

    \bottomrule
    \end{tabular}%
    }

    \global\softtabdatafalse

    \par}

    \caption{
    Pass@\(K\) results (\(\times 100\)) on mathematical benchmarks in the Hard Setting.
    Underlining marks the best Base/GRPO/ES result in each task cell; bold and shading mark the best and second-best Average cells across all five methods.
    }
    \label{tab:rq1-deepseek-math}
    \vspace{-10pt}
\end{table*}

\begin{figure}[H]
    \centering
    \includegraphics[width=0.47\textwidth]{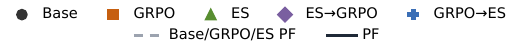}\par
    \vspace{1pt}
    \begin{subfigure}[t]{0.32\textwidth}
        \centering
        \includegraphics[width=\linewidth]{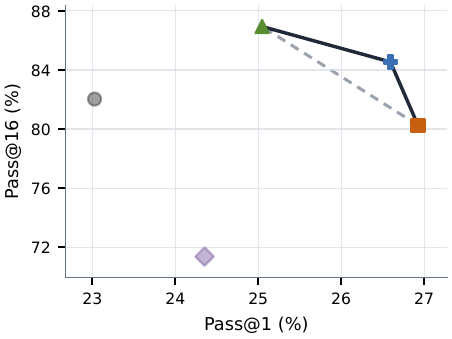}
        \caption{\textbf{GPQA}\\[1pt]
        \subfigcontext{Llama-3.2-3B-Instruct}}
        \label{fig:rq1-pareto-llama-gpqa}
    \end{subfigure}\hfill
    \begin{subfigure}[t]{0.32\textwidth}
        \centering
        \includegraphics[width=\linewidth]{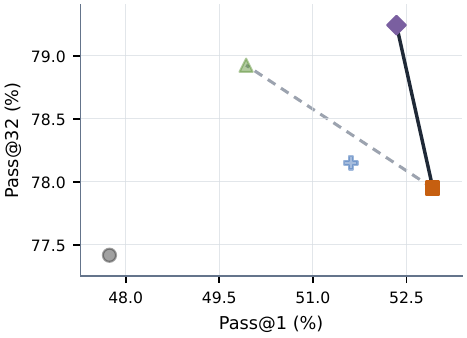}
        \caption{\textbf{Math Average}\\[1pt]
        \subfigcontext{DeepSeek-R1-Distill-Qwen-1.5B}}
        \label{fig:rq1-pareto-deepseek-math-average}
    \end{subfigure}\hfill
    \begin{subfigure}[t]{0.32\textwidth}
        \centering
        \includegraphics[width=\linewidth]{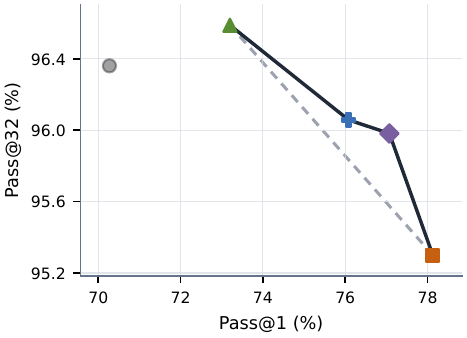}
        \caption{\textbf{GSM8K}\\[1pt]
        \subfigcontext{Qwen2.5-1.5B-Instruct}}
        \label{fig:rq1-pareto-qwen15-gsm8k}
    \end{subfigure}
    \caption{
    Representative Pass@1--Pass@\(K\) Pareto fronts across models and tasks.
    The math average is computed equally over AIME24, AIME25, AMC23, and MATH-500; higher values are better on both axes.
    Gray dashed lines connect the reference Pareto fronts formed by Base, GRPO, and ES.
    Solid dark lines trace the full Pareto fronts after adding the endpoints from the two sequential compositions.
    Faded markers are dominated in the corresponding pairwise comparison.
    The two sequential compositions add non-dominated trade-off points in all three settings.
    }
    \label{fig:rq1-deepseek-math-pareto}
\end{figure}

\textbf{Pareto Trade-offs from Sequential Compositions.}
Figure~\ref{fig:rq1-deepseek-math-pareto} treats Pass@1 and Pass@\(K\) as paired objectives in three representative model--task settings.
In each panel, the reference Pareto front formed by Base, GRPO, and ES reflects the trade-off between GRPO's higher Pass@1 and ES's higher large-\(K\) coverage.
Adding the endpoints from the two sequential compositions introduces additional non-dominated points.
In particular, ES$\rightarrow$GRPO attains the highest Pass@32 on the Hard-Setting math average while retaining most of GRPO's Pass@1 gain, and GRPO$\rightarrow$ES supplies an intermediate trade-off on GPQA with Llama-3.2-3B-Instruct.

\obsbox{
\textbf{Takeaways.}
\begin{itemize}[leftmargin=*]
    \item \textbf{Sampling across ES perturbations can increase Pass@\(K\).} ES perturbations produce population members with different response distributions and success rates. Sampling once from each member is at least as likely to include a correct answer as drawing the same number of responses from a single policy with the same average success rate. Under the conditions in Section~\ref{sec:rq1-theoretical-results}, the ES update can preserve this advantage in the updated model.
    \item \textbf{ES improves Pass@1 and attains higher Pass@\(K\) than GRPO.} Across the Easy and Hard Settings, ES improves average Pass@1, Pass@16, and Pass@32 over the Base Model and exceeds GRPO on average Pass@16 and Pass@32. Representative training dynamics further show that ES largely avoids the entropy decline and large-\(K\) degradation observed under GRPO.
    \item \textbf{The two sequential compositions add new Pass@1--Pass@\(K\) trade-offs.} Under the same total update budget, adding the endpoints from ES$\rightarrow$GRPO and GRPO$\rightarrow$ES expands the Pareto fronts in all three representative settings and provides additional non-dominated points.
\end{itemize}
}

\section{RQ2: Does ES Necessarily Cause Catastrophic Forgetting?}

RQ2 examines the scale and distribution of ES-induced parameter changes, identifies which updates account for performance gains, and evaluates whether large parameter drift necessarily leads to catastrophic forgetting.

\subsection{ES Induces Greater Parameter Drift}

ES evaluates full-parameter perturbations and can accumulate substantial whole-model parameter drift over successive updates. Prior works identify this large parameter movement as a characteristic difference between ES and gradient-based post-training \citep{hoy2026matching,abdi2026catastrophic}. We first quantify this movement relative to matched GRPO training.

\FloatBarrier

\begin{table*}[t]
    \centering
    {\fontsize{8}{10}\selectfont
    
    \renewcommand{\arraystretch}{1.10}
    \setlength{\tabcolsep}{3.5pt}
    \begin{tabular*}{\textwidth}{@{\extracolsep{\fill}}lccccc@{}}
    \toprule
    \multirow[c]{2}{*}{Model}
    & \multirow[c]{2}{*}{GRPO}
    & \multirow[c]{2}{*}{Full ES}
    & \multicolumn{3}{c}{Magnitude-Thresholded ES} \\
    \cmidrule(lr){4-6}
    & &
    & $\tau=1.0\times10^{-3}$
    & $\tau=1.5\times10^{-3}$
    & $\tau=2.0\times10^{-3}$ \\
    \midrule
    Qwen2.5-1.5B-Instruct
    & 0.0475 & 1.933 (40.7$\times$)
    & 1.615 (79.11\%)
    & 1.242 (92.47\%)
    & 0.871 (\textbf{97.57\%}) \\
    Llama-3.2-3B-Instruct
    & 0.0954 & 4.185 (43.9$\times$)
    & 3.482 (78.08\%)
    & 2.589 (92.64\%)
    & 1.707 (\textbf{97.89\%}) \\
    Qwen2.5-7B-Instruct
    & 0.0831 & 3.664 (44.1$\times$)
    & 3.032 (78.41\%)
    & 2.218 (93.02\%)
    & 1.427 (\textbf{98.11\%}) \\
    DeepSeek-R1-Distill-Qwen-1.5B
    & 0.0548
    & 2.338 (42.7$\times$)
    & 2.201 (62.18\%)
    & 2.017 (77.61\%)
    & 1.772 (\textbf{87.29\%}) \\
    \bottomrule
    \end{tabular*}
    \par}
    \caption{Relative whole-model \(L_2\) distance (\(\times10^{-2}\)) and the distribution of nonzero ES updates across magnitude thresholds. Parentheses in the Full ES column report the distance relative to GRPO; parentheses in the magnitude-thresholded columns report \(s_\tau\), the percentage of nonzero updates with magnitudes in \((0,\tau]\).}
    \label{tab:rq2-parameter-drift}
\end{table*}
Following prior analyses of parameter drift of ES \citep{hoy2026matching,abdi2026catastrophic}, we measure whole-model movement using relative $L_2$ distance,
\begin{equation}
D_{\mathrm{rel}}(\theta,\theta_0)
=
\frac{\lVert\theta-\theta_0\rVert_2}{\lVert\theta_0\rVert_2}.
\label{eq:rq2-parameter-drift}
\end{equation}
Here, $\theta_0$ and $\theta$ denote the parameters before and after post-training, respectively.
% Normalization by $\lVert\theta_0\rVert_2$ enables comparisons across model sizes.
Table~\ref{tab:rq2-parameter-drift} shows that Full ES is 40.7--44.1 times
farther from initialization than GRPO across the four models, confirming the substantially different parameter geometry reported in prior work.
\citet{hoy2026matching} explain this movement by decomposing ES dynamics into reward-aligned progress and diffusion along locally flat or weakly reward-relevant directions.
Along these directions, the expected optimization signal is small, but stochastic perturbation aggregation can accumulate as a high-dimensional random walk.
Thus, the large total drift of ES may partly reflect accumulated movement in many directions with little effect on reward.

\subsection{ES Effects Are Concentrated in a Small Subset of Larger-Magnitude Updates}

However, relative whole-model distance aggregates changes across all parameter coordinates and does not reveal their magnitude distribution.
We therefore quantify the proportions of parameter changes within and beyond a range of magnitude thresholds.

Let $\Delta\theta_i=\theta_i-\theta_{0,i}$ denote the change of parameter $i$ from the Base Model.
For a threshold $\tau$, we first quantify the fraction of nonzero updates whose magnitudes fall within $(0,\tau]$:
\begin{equation}
s_\tau
=
\frac{\left|\left\{i:0<|\Delta\theta_i|\leq\tau\right\}\right|}
     {\left|\left\{i:|\Delta\theta_i|>0\right\}\right|},
\label{eq:rq2-update-sparsity}
\end{equation}
The complementary fraction, $1-s_\tau$, consists of updates with magnitudes greater than $\tau$.
Table~\ref{tab:rq2-parameter-drift} shows that most changed parameter coordinates have relatively small update magnitudes, whereas larger-magnitude updates form a smaller subset.
For example, at $\tau=1.5\times10^{-3}$, which falls within the magnitude range of a single ES update, the interval $(0,\tau]$ contains 77.6--93.0\% of the nonzero updates across the four models, leaving 7.0--22.4\% above the threshold.

\begin{figure}[!t]
    \centering
    \includegraphics[width=0.50\linewidth]{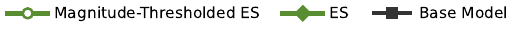}\par
    \vspace{1pt}
    \begin{subfigure}[t]{0.48\linewidth}
        \centering
        \includegraphics[width=\linewidth]{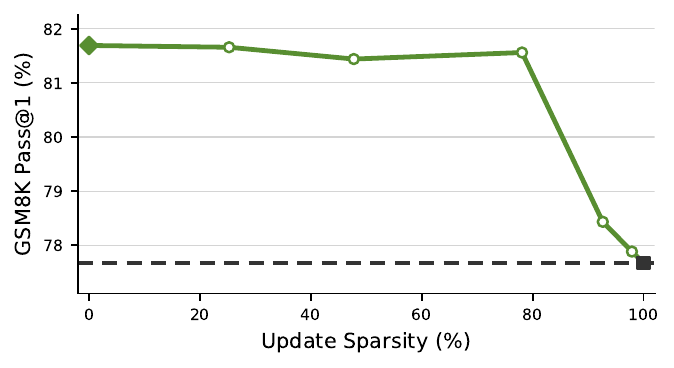}
        \caption{\textbf{GSM8K}\\[1pt]
        \subfigcontext{Llama-3.2-3B-Instruct}}
        \label{fig:rq2-update-sparsity-llama}
    \end{subfigure}\hfill
    \begin{subfigure}[t]{0.48\linewidth}
        \centering
        \includegraphics[width=\linewidth]{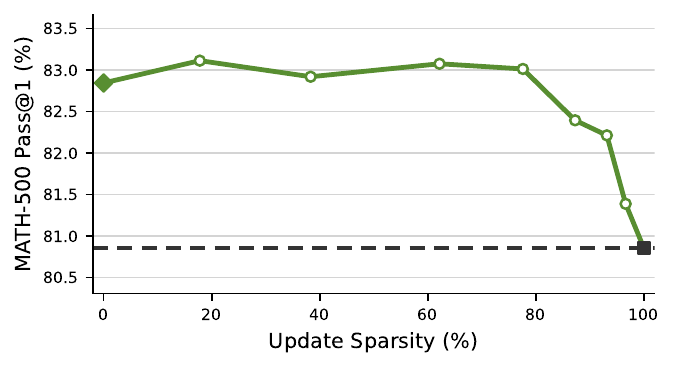}
        \caption{\textbf{MATH-500}\\[1pt]
        \subfigcontext{DeepSeek-R1-Distill-Qwen-1.5B}}
        \label{fig:rq2-update-sparsity-deepseek}
    \end{subfigure}
    \caption{Target-task Pass@1 across update-sparsity levels in the Easy and Hard Settings. The dashed lines indicate
    previously recorded Base Model performance, which also forms the connected
    endpoint at 100\% update sparsity.}
    \label{fig:rq2-update-sparsity-accuracy}
\end{figure}

Having established that larger-magnitude updates form a smaller subset, we next test whether removing the more numerous small-magnitude updates preserves task performance.
For each $\tau$, we construct a magnitude-thresholded checkpoint by setting $\Delta\theta_i=0$ for coordinates satisfying $0<|\Delta\theta_i|\leq\tau$.
Under this operation, $s_\tau$ is the Update Sparsity, and Full ES corresponds to $s_\tau=0$.
Figure~\ref{fig:rq2-update-sparsity-accuracy} shows that target-task Pass@1
remains broadly stable as progressively more small-magnitude updates are set to zero,
with noticeable degradation emerging only at high update sparsity.
Together, the update-magnitude statistics and thresholding experiments reveal magnitude sparsity with corresponding functional concentration: despite perturbing all parameters, ES produces larger-magnitude updates in a limited parameter subset, and retaining these updates preserves most of the performance gains.
The retained coordinates can therefore be viewed as defining an \textbf{approximately performance-preserving coordinate subspace} identified after training.

\paragraph{Our Explanation for Why ES can Work on Full-Parameter LLM}

In our view, this \textbf{post-hoc structure} helps explain why full-parameter ES can remain effective in the high-dimensional parameter space of an LLM, because its performance gains do not depend equally on updates throughout the entire space. \citet{frankle2018lottery} mentioned that larger models contain more such effective structures, so ES can work on LLMs with billions of parameters.

We further compare the locations and scales of the largest ES and GRPO updates. The largest ES updates occur
mainly in LayerNorm weights and attention
projections. In Llama-3.2-3B-Instruct, the maximum magnitude is \(0.01171875\):
five of the nine maximum coordinates are input-LayerNorm weights, while the remaining maxima include attention and MLP projections.
In DeepSeek-R1-Distill-Qwen-1.5B, the maximum is \(0.015625\); 117 of its 144
maximum coordinates are LayerNorm weights and 24 are attention-projection
biases. Normalization parameters further account for 72 and 80 of the 100
largest updates in
Llama-3.2-3B-Instruct and DeepSeek-R1-Distill-Qwen-1.5B, respectively.
By contrast, under GRPO, the
largest Llama-3.2-3B-Instruct update is \(0.00024414\) (\(48\times\) smaller)
and its top 100 all occur in token embeddings; the
DeepSeek-R1-Distill-Qwen-1.5B GRPO endpoint peaks at \(0.00097656\)
(\(16\times\) smaller), with its top 100 all in the language-model head. This
contrast suggests that ES may adapt through normalization and attention
parameters: LayerNorm rescales hidden states, while
attention projections route information. GRPO instead concentrates its largest changes in token
embeddings and the language-model head, adjacent to input
representations and output logits, consistent with token-level
gradient optimization.

\subsection{Large Parameter Drift Does Not Necessarily Indicate Broad Forgetting}

We next test whether the larger parameter movement of ES leads to held-out performance degradation.
Tables~\ref{tab:rq1-instruct-final} and~\ref{tab:rq2-deepseek-nonmath} report held-out-task performance for the evaluated models in the Easy and Hard Settings, respectively.
Table~\ref{tab:rq2-heldout-change} in Appendix~\ref{app:heldout-performance-changes} additionally reports metric-wise performance changes for each Easy Setting model.

\begin{table}[t]
    \centering
    \vspace{-15pt}
    {\fontsize{8}{10}\selectfont
    
    \setlength{\tabcolsep}{0.8pt}
    \global\softtabdatafalse
    \begin{tabular*}{\textwidth}{@{\extracolsep{\fill}}l*{15}{N}@{}}
    \multicolumn{16}{c}{\textbf{Pass Metrics}} \\
    \toprule
    & \multicolumn{3}{c}{GPQA}
    & \multicolumn{3}{c}{MBPP}
    & \multicolumn{3}{c}{CSQA}
    & \multicolumn{3}{c}{Countdown}
    & \multicolumn{3}{c}{Average} \\
    \cmidrule(l{1pt}r{1pt}){2-4}\cmidrule(l{1pt}r{1pt}){5-7}
    \cmidrule(l{1pt}r{1pt}){8-10}\cmidrule(l{1pt}r{1pt}){11-13}
    \cmidrule(l{1pt}r{1pt}){14-16}
    Method & @1 & @16 & @32
    & @1 & @16 & @32
    & @1 & @16 & @32
    & @1 & @16 & @32
    & @1 & @16 & @32 \\
    \midrule
    \multicolumn{16}{l}{\textbf{DeepSeek-R1-Distill-Qwen-1.5B}} \\
    \global\softtabdatatrue Base & 24.2 & 80.8 & 91.4 & 62.9 & 85.6 & 87.6 & 46.8 & \underline{89.2} & \underline{93.7} & 37.5 & \underline{79.9} & \underline{85.7} & 42.9 & 83.9 & 89.6 \\
    \grpomethod & \underline{31.8} & \underline{86.4} & 91.9 & \underline{66.6} & \underline{87.0} & \underline{88.3} & \underline{47.5} & 88.9 & 93.6 & 36.6 & 77.5 & 82.2 & \textbf{45.6} & 85.0 & 89.0 \\
    \esmethod & 28.0 & 84.6 & \underline{92.4} & 63.5 & 85.9 & 87.2 & 46.1 & 88.9 & 93.4 & \underline{38.5} & 79.4 & 84.4 & 44.0 & 84.7 & 89.3 \\
    \esmethod$\rightarrow$\grpomethod & 30.6 & 84.5 & 91.4 & 64.6 & 86.7 & 87.9 & 46.8 & 88.8 & 93.3 & 38.8 & 80.6 & 86.8 & \cellcolor{gray!35}45.2 & \cellcolor{gray!35}85.1 & \textbf{89.9} \\
    \grpomethod$\rightarrow$\esmethod & 30.6 & 86.7 & 93.4 & 64.2 & 85.2 & 87.2 & 47.3 & 88.7 & 93.2 & 38.1 & 79.9 & 85.3 & 45.1 & \textbf{85.1} & \cellcolor{gray!35}89.8 \\
    \bottomrule
    \end{tabular*}
    \global\softtabdatafalse

    \par}
    \caption{Pass@\(K\) results (\(\times 100\)) on four held-out benchmarks in the one-epoch Hard Setting. Underlining marks the best Base/GRPO/ES result in each task cell; bold and shading mark the best and second-best Average cells across all five methods.}
    \label{tab:rq2-deepseek-nonmath}
    \vspace{-15pt}
\end{table}

For all three Easy Setting models, the Pass@32 change averaged over the
five held-out tasks is positive under ES but negative under GRPO
(Table~\ref{tab:rq2-heldout-change}).
In the Hard Setting, both methods improve average Pass@1 and Pass@16 on the held-out non-mathematical benchmarks, while ES achieves higher average Pass@32 than GRPO (Table~\ref{tab:rq2-deepseek-nonmath}).
The corresponding majority-vote results show that ES exceeds both the Base Model and GRPO in average Maj@16 and Maj@32 (Table~\ref{tab:app-rq2-deepseek-nonmath-majority}).

Taken together, these results suggest that ES generally maintains held-out performance despite large parameter drift.
This finding differs from \citet{abdi2026catastrophic}, who report catastrophic forgetting under ES.
Their evidence was limited to a small training set and a single model, training task, and held-out benchmark, making broad capability forgetting difficult to distinguish from overfitting to the training set.

However, performance drops on some tasks are especially consequential in practically relevant continual-learning scenarios, where deployed models must repeatedly adapt to new tasks and knowledge while reliably preserving prior capabilities.
Prior exploratory work suggests that the effects of ES on prior capabilities may depend on the task and training configuration \citep{hoy2026matching}.
This raises concern that ES may preserve capabilities inconsistently across tasks and repeated updates.
\obsbox{
\textbf{Takeaways.}
\begin{itemize}[leftmargin=*]
    \item \textbf{ES induces greater whole-model parameter drift than GRPO.} ES-trained models move substantially farther from their initial parameters than their GRPO-trained counterparts.
    \item \textbf{ES updates exhibit magnitude sparsity.} Larger-magnitude updates occupy a small subset of parameters, while ignoring smaller updates largely preserves task performance. These larger updates occur mainly in normalization and attention parameters.
    \item \textbf{Large parameter drift does not necessarily lead to catastrophic forgetting.} ES generally preserves held-out performance and performs better than GRPO across the Easy and Hard Settings.
\end{itemize}
}

\section{RQ3: What Parameter Settings and Estimators Make ES Effective and Scalable?}

RQ3 examines how reward normalization, perturbation scale, population size, and estimator choice affect effective and scalable ES training.

\subsection{Reward Normalization, Perturbation Scale, and Population-Size Scaling}

\input{figures/rq3_population_scaling_checkpoints.tex}
\begin{table}[H]
    \centering
    {\fontsize{8}{9.5}\selectfont
    
    \setlength{\tabcolsep}{2pt}
    \begin{tabular*}{\textwidth}{@{\extracolsep{\fill}}l*{12}{c}@{}}
        \toprule
        & \multicolumn{4}{c}{\textbf{Qwen2.5-0.5B-Instruct}}
        & \multicolumn{4}{c}{\textbf{Qwen2.5-1.5B-Instruct}}
        & \multicolumn{4}{c}{\textbf{Qwen2.5-3B-Instruct}} \\
        \cmidrule(lr){2-5}\cmidrule(lr){6-9}\cmidrule(lr){10-13}
        Update
        & \(N=8\) & \(N=16\) & \(N=32\) & \(N=64\) (ref.)
        & \(N=8\) & \(N=16\) & \(N=32\) & \(N=64\) (ref.)
        & \(N=8\) & \(N=16\) & \(N=32\) & \(N=64\) (ref.) \\
        \midrule
        \rqPopulationCheckpointRows
        \bottomrule
    \end{tabular*}
    \par}
    \caption{Smoothed GSM8K rewards across model and population sizes. Shading marks reduced-population results within \(0.01\) of the \(N=64\) reference; values use debiased exponential smoothing with weight \(0.99\).}
    \label{tab:rq3-population-checkpoints}
    \vspace{-10pt}
\end{table}

\paragraph{Reward Normalization.}
We z-score population rewards before using them to weight perturbation directions, making each update depend on relative rather than absolute reward scales.
In the matched single-point ES ablation, z-score normalization yields higher mean rewards than no normalization throughout the evaluated updates after initialization, indicating improved reward-guided optimization in this setting (Figure~\ref{fig:rq123-overview}(c), item~(1)).

\paragraph{Perturbation Scale.}
The perturbation scale \(\sigma\) sets both the radius of parameter-space exploration and the smoothing level of the effective objective \(F_\sigma\).
When \(\sigma\) is too small, ES explores a narrow neighborhood and may overfit the observed rewards, causing the search to become trapped around a local optimum.
When \(\sigma\) is too large, perturbations span an overly broad region, leading to excessive exploration that can destabilize training and collapse performance.
During training, \(\sigma\) should therefore be selected to limit early reward overfitting while maintaining stable reward-guided progress.

\paragraph{Population-Size Scaling.}
Section~4.2 shows that ES can preserve task performance with a small subset of larger-magnitude updates.
Larger models may contain more such subsets with similar effects, making task-improving perturbations denser around pretrained weights, consistent with \citet{frankle2018lottery,gan2026neuralthickets}.
We therefore compare matched GSM8K runs of Qwen2.5-\(\{0.5,1.5,3\}\)B-Instruct with \(N\in\{8,16,32,64\}\) to test whether fewer directions suffice at larger scales.
At update 300, Table~\ref{tab:rq3-population-checkpoints} shows that only \(N=32\) is within \(0.01\) of \(N=64\) for 0.5B, while both \(N=16\) and \(N=32\) meet this threshold for 1.5B and 3B; correspondingly, the \(N=16\)-to-\(N=64\) gap falls from \(0.0352\) to \(0.0051\) and \(0.0030\).
Together with the trajectories in Appendix~\ref{app:population-scaling-trajectories}, these results indicate that smaller populations approach the \(N=64\) reward as model scale grows.
Accordingly, the population required for stable ES training appears to decrease with model scale.

\subsection{Why Two-Point ZO Intuition Does Not Directly Transfer to ES for Reasoning}

Zeroth-order (ZO) optimization is a popular gradient-free approach that typically estimates gradients by subtracting objective values at two symmetric perturbations and is commonly evaluated on non-reasoning NLP tasks such as SST-2~\citep{malladi2023mezo}.
ES can instead use one evaluation per direction, as in our one-point implementation, or the same antithetic two-point estimator; under matched objectives and perturbations, antithetic ES is algebraically identical to two-point ZO.
The benefit of subtraction depends on evaluation: supervised tasks can reuse fixed data, whereas reasoning tasks regenerate autoregressive responses whose early-token divergence can weaken paired covariance.
In our matched GSM8K run, two-point ES provides no training-reward or held-out advantage (Figure~\ref{fig:rq123-overview}(c), item~(3)), while Appendix~\ref{app:estimator-boundary} finds raw variance reduction for SST-2 but not reliably for regenerated GSM8K rewards.
Thus, two-point ZO gains on non-reasoning tasks do not necessarily transfer to ES for reasoning; we favor one-point estimation when coupling is weak.

\obsbox{
\textbf{Takeaways.}
\begin{itemize}[leftmargin=*]
    \item \textbf{Z-score normalization yields higher training rewards.} In the matched one-point ES ablation, z-score normalization consistently outperforms no normalization after initialization.
    \item \textbf{Larger models can use smaller ES populations.} As model scale increases, reduced-population runs more closely approach the \(N=64\) reference. At update 300, \(N=16\) is within \(0.01\) of \(N=64\) for the 1.5B and 3B models, whereas only \(N=32\) meets this criterion for the 0.5B model.
    \item \textbf{Two-point estimation provides no advantage over one-point estimation.} In the matched GSM8K experiment, two-point ES improves neither training reward nor held-out performance.
\end{itemize}
}

\section{Conclusion and Future Work}

\paragraph{Conclusion.}
This work examines ES's reasoning ability gains, possible forgetting as a side effect, and conditions for effective training for LLMs.
Theoretically and empirically, we demonstrate that ES improves Pass@1 while maintaining broader Pass@\(K\) coverage than GRPO, which can exhibit entropy collapse alongside a decline in Pass@\(K\).
Held-out evaluations further indicate that large parameter movement does not necessarily erase existing capabilities, in contrast to previous reports.
Our design study identifies effective parameter and estimator choices and finds that larger models can train stably with smaller populations.
Together, these results position ES as a distinct reasoning post-training paradigm rather than a memory-efficient GRPO alternative.

\paragraph{Future Work.}
Future work should better exploit ES's incentivized-reasoning advantage by expanding access to correct paths assigned low probability by the base policy.
Continual learning should be further explored to clarify how parameter drift affects previously acquired capabilities over longer training horizons spanning multiple tasks, and to investigate whether approaches such as parameter-efficient updates can suppress harmful drift without constraining beneficial exploration.
\bibliographystyle{assets/plainnat}
\bibliography{paper}

\clearpage
\appendix
\section{Proofs for the RQ1 Diversity--Coverage Analysis}
\label{app:rq1-proofs}

This appendix provides the auxiliary identities, qualifications, and proofs behind Lemmas~\ref{lem:rq1-local-policy-js}--\ref{lem:rq1-selection-gain} and Proposition~\ref{prop:rq1-center-transfer}.
The Jensen--Shannon (JS) definitions and entropy representation follow the standard formulation of \citet{lin1991divergence}.
We use the usual information-theoretic forms of the data-processing and Pinsker inequalities~\citep{cover2006elements}; the arithmetic--geometric mean, Taylor, and Jensen inequalities are invoked explicitly at the steps where they are used.

\subsection{Proof of Lemma~\ref{lem:rq1-local-policy-js}}
\label{app:proof-local-policy-js}

Finite response-distribution entropies imply this identity:
\begin{equation}
H(\bar\pi_N)
=
\frac{1}{N}\sum_{i=1}^N H(\pi_i)
+
\operatorname{JS}^{\mathrm{pol}}_N(x).
\label{eq:app-policy-entropy-decomposition}
\end{equation}
Thus, $\operatorname{JS}^{\mathrm{pol}}_N$ measures diversity across population members rather than sampling entropy within one policy.
For a GRPO rollout group drawn from one behavior policy, all member policies are identical and the corresponding cross-parameter JS is zero, although within-policy sampling entropy can remain nonzero.

Take $N$ and the parameter dimension to be fixed and finite, and let $\mathcal{Y}_x$ be the countable response space for prompt $x$.
Write $q_\Delta(y)=\pi_{\theta+\Delta}(y\mid x)$, $q_0=q_{\Delta=0}$,
$s_\theta=s_\theta(y\mid x)$, $\bar\Delta=N^{-1}\sum_i\Delta_i$,
$\bar q=N^{-1}\sum_iq_{\Delta_i}$, and
$\lVert\Delta_{1:N}\rVert=\max_i\lVert\Delta_i\rVert$.
For some $r>0$, assume that $q_\Delta$ has common positive support and finite entropy on $\{\lVert\Delta\rVert\le r\}$, is four times continuously differentiable in $\Delta$, and has finite Fisher information at $\Delta=0$.
Define the per-response JS integrand
$\ell_y(\Delta_{1:N})=N^{-1}\sum_iq_{\Delta_i}(y)\log[q_{\Delta_i}(y)/\bar q(y)]$.
For every multi-index $|\nu|\le4$, assume
$\sup_{\lVert\Delta_{1:N}\rVert\le r}|\partial^\nu\ell_y(\Delta_{1:N})|
\le M_\nu(y)$ for an envelope satisfying
$\sum_{y\in\mathcal{Y}_x}M_\nu(y)<\infty$.
These conditions justify termwise differentiation of the response sum and a uniform fourth-order Taylor remainder.
For local offsets $\Delta_1,\ldots,\Delta_N$, Taylor expansion gives
$q_\Delta=q_0[1+s_\theta^\top\Delta]+O(\lVert\Delta\rVert^2)$ and
$\bar q=q_0[1+s_\theta^\top\bar\Delta]+O(\lVert\Delta_{1:N}\rVert^2)$.
Substituting into the KL divergence and using
$\mathbb{E}_{q_0}[s_\theta]=0$ and
$\mathbb{E}_{q_0}[s_\theta s_\theta^\top]=\mathcal{I}_x(\theta)$ yields the quadratic term below.
Retaining the third- and fourth-order terms gives
\begin{equation}
\begin{gathered}
\operatorname{JS}^{\mathrm{pol}}_N(x)
=
\frac{1}{2N}\sum_{i=1}^N
(\Delta_i-\bar\Delta)^\top
\mathcal{I}_x(\theta)
(\Delta_i-\bar\Delta)
\\[-0.2ex]
{}+\mathcal{T}_3(\Delta_{1:N})
+\mathcal{R}_4(\Delta_{1:N}),
\end{gathered}
\label{eq:app-local-js-expansion}
\end{equation}
where $\mathcal{T}_3$ is homogeneous of degree three and
$|\mathcal{R}_4(\Delta_{1:N})|\le C\lVert\Delta_{1:N}\rVert^4$
whenever $\lVert\Delta_{1:N}\rVert\le r$.
For $\Delta_i=\sigma\epsilon_i$, the covariance becomes
\begin{equation}
\mathbb{E}
\left[
(\Delta_i-\bar\Delta)(\Delta_i-\bar\Delta)^\top
\right]
=
\sigma^2\left(1-\frac{1}{N}\right)I.
\label{eq:app-centered-perturbation-covariance}
\end{equation}
Therefore, the expected quadratic term in Equation~\eqref{eq:app-local-js-expansion} is
\begin{equation}
\frac{\sigma^2}{2}
\left(1-\frac{1}{N}\right)
\operatorname{tr}\mathcal{I}_x(\theta).
\end{equation}
Let $E_\sigma=\{\max_i\lVert\sigma\epsilon_i\rVert\le r\}$.
It is sign invariant, so
$\mathbb{E}[\mathcal{T}_3(\sigma\epsilon_{1:N})\mathbf{1}_{E_\sigma}]=0$.
The remainder on $E_\sigma$ has expectation $O(\sigma^4)$.
Moreover, $\operatorname{JS}^{\mathrm{pol}}_N\le\log N$ and
$\Pr(E_\sigma^c)=O(e^{-c/\sigma^2})$, so the JS tail and quadratic truncation are $o(\sigma^4)$, establishing the expansion in Equation~\eqref{eq:rq1-local-policy-js}.

\subsection{Proof of Lemma~\ref{lem:rq1-population-coverage}}
\label{app:proof-population-coverage}

The correctness verifier deterministically maps each response distribution $\pi_i(\cdot\mid x)$ to
$B_i(x)=\operatorname{Bernoulli}(p_i(x))$.
The data-processing inequality for generalized JS divergence therefore proves Equation~\eqref{eq:rq1-js-data-processing}.
This projection is necessary because full policy JS may arise solely from variation among incorrect responses.

For a fixed prompt $x$, suppress the dependence on $x$ and write $p_i=p_i(x)$ and $\bar p=N^{-1}\sum_i p_i$.
By the arithmetic--geometric mean inequality,
\begin{equation}
\prod_{i=1}^N(1-p_i)
\le
\left(\frac{1}{N}\sum_{i=1}^N(1-p_i)\right)^N
=
(1-\bar p)^N.
\label{eq:app-coverage-amgm}
\end{equation}
Subtracting both sides from one proves
$P_N^{\mathrm{pop}}\ge P_N^{\mathrm{same}}$.
Equality in the arithmetic--geometric mean inequality holds exactly when
$1-p_1=\cdots=1-p_N$, equivalently $p_1=\cdots=p_N$.

For the local relation, let $p_i=\bar p+\xi_i$, $\sum_i\xi_i=0$, and $\beta=1-\bar p$, where $\xi_i$ is the centered member-success deviation and $\beta$ the average failure probability.
Assume that $\bar p$ remains in a compact subset of $(0,1)$.
At $\xi=0$, expand the product:
\begin{equation}
\prod_{i=1}^N(\beta-\xi_i)
=
\beta^N-\frac{\beta^{N-2}}{2}\sum_{i=1}^N\xi_i^2
+O(\lVert\xi\rVert^3),
\label{eq:app-failure-product-expansion}
\end{equation}
Subtracting from one gives the coverage difference:
\begin{equation}
P_N^{\mathrm{pop}}-P_N^{\mathrm{same}}
=
\frac{\beta^{N-2}}{2}\sum_{i=1}^N\xi_i^2
+O(\lVert\xi\rVert^3).
\label{eq:app-coverage-variance}
\end{equation}
For $h(p)=-p\log p-(1-p)\log(1-p)$, the binary entropy satisfies $h''(p)=-1/[p(1-p)]$.
A Taylor expansion of the success-JS definition in Lemma~\ref{lem:rq1-population-coverage}, with the linear term canceled by $\sum_i\xi_i=0$, yields
\begin{equation}
\operatorname{JS}^{\mathrm{succ}}_N
=
\frac{1}{2N\bar p(1-\bar p)}
\sum_{i=1}^N\xi_i^2
+O(\lVert\xi\rVert^3).
\label{eq:app-success-js-variance}
\end{equation}
Eliminating $\sum_i\xi_i^2$ between Equations~\eqref{eq:app-coverage-variance} and~\eqref{eq:app-success-js-variance} gives
\begin{equation}
P_N^{\mathrm{pop}}(x)-P_N^{\mathrm{same}}(x)
=
N\bar p(x)(1-\bar p(x))^{N-1}
\operatorname{JS}^{\mathrm{succ}}_N(x)
{}\,+
O(\lVert\xi(x)\rVert^3).
\label{eq:rq1-coverage-js-link}
\end{equation}

\subsection{Proof and Illustration of Lemma~\ref{lem:rq1-selection-gain}}
\label{app:proof-selection-expansion}

Under a uniform draw of member index $i$, $\mathbb{E}_i[w_i]=1/N$ and $\mathbb{E}_i[p_i]=\bar p$. Hence
\begin{equation}
N\operatorname{Cov}_i(w_i,p_i)
=
\sum_{i=1}^N w_i p_i-\bar p
=
p_w-\bar p,
\end{equation}
which proves the covariance identity in Lemma~\ref{lem:rq1-selection-gain}.

For intuition only, define the prompt-adaptive oracle gate
\begin{equation}
\begin{gathered}
\widetilde w_i(x)
=
\frac{\exp(\lambda p_i(x))}{\sum_{j=1}^N\exp(\lambda p_j(x))}, \\
\widetilde p_w(x) = \sum_{i=1}^N\widetilde w_i(x)p_i(x),
\end{gathered}
\label{eq:rq1-oracle-gate}
\end{equation}
where $\lambda>0$.
For fixed $N\ge2$, let $p_i=\bar p+\xi_i$ and assume that $\bar p(x)$ is bounded away from zero and one.
This parameterization gives the following normalized weights:
\begin{equation}
\widetilde w_i
=
\frac{\exp(\lambda\xi_i)}
{\sum_{j=1}^N\exp(\lambda\xi_j)}.
\end{equation}
Since $\sum_i\xi_i=0$, the ratio expands as
\begin{equation}
\widetilde w_i
=
\frac{1}{N}+\frac{\lambda}{N}\xi_i
+O(\lVert\xi\rVert^2).
\label{eq:app-softmax-weight-expansion}
\end{equation}
This expansion gives the selected-policy shift:
\begin{equation}
\widetilde p_w-\bar p
=
\sum_{i=1}^N \widetilde w_i\xi_i
=
\frac{\lambda}{N}\sum_{i=1}^N\xi_i^2
+O(\lVert\xi\rVert^3).
\label{eq:app-selection-variance}
\end{equation}
Substituting Equation~\eqref{eq:app-success-js-variance} into Equation~\eqref{eq:app-selection-variance} gives
\begin{equation}
\widetilde p_w(x)-\bar p(x)
=
2\lambda\bar p(x)(1-\bar p(x))
\operatorname{JS}^{\mathrm{succ}}_N(x)
+O(\lVert\xi(x)\rVert^3).
\label{eq:rq1-selection-js-link}
\end{equation}
This calculation illustrates how success/failure predictive JS becomes exploitable when fitness is calibrated to success.

\subsection{Proof of Proposition~\ref{prop:rq1-center-transfer}}
\label{app:proof-center-transfer}

Data processing yields the relaxed full-policy KL bound:
\begin{equation}
D_{\mathrm{KL}}(B_w\,\|\,B_{\theta^+})
\le
D_{\mathrm{KL}}(\pi_w\,\|\,\pi_{\theta^+}).
\end{equation}
It allows the center policy to use different responses while preserving total verifier success probability.
This empirically testable condition connects the analytical comparator to the actual ES center.

Define the prompt-wise transfer divergence as
\begin{equation}
D_x
=
D_{\mathrm{KL}}
\left(B_w(x)\,\|\,B_{\theta^+}(x)\right).
\end{equation}
For Bernoulli distributions, their total variation distance is the absolute difference between their success probabilities.
Pinsker's inequality bounds this difference:
\begin{equation}
\left|p_w(x)-p_{\theta^+}(x)\right|
\le
\sqrt{\frac{D_x}{2}}.
\label{eq:app-binary-pinsker}
\end{equation}
For the $K$-sample success map $f_K(p)=1-(1-p)^K$, the derivative is bounded as follows:
\begin{equation}
0\le f_K'(p)=K(1-p)^{K-1}\le K
\qquad\text{for }p\in[0,1],
\end{equation}
Thus, $f_K$ is $K$-Lipschitz.
From Equation~\eqref{eq:app-binary-pinsker}, Lipschitz continuity implies
\begin{equation}
f_K(p_{\theta^+}(x))
\ge
f_K(p_w(x))-K\sqrt{\frac{D_x}{2}}.
\end{equation}
Taking expectation over $x$ and applying Jensen's inequality to the concave square-root function gives
\begin{gather}
J_K(\pi_{\theta^+})
\ge
J_K(\pi_w)-K\mathbb{E}_{x\sim\mathcal D}\sqrt{\frac{D_x}{2}}
\\
J_K(\pi_{\theta^+})
\ge
J_K(\pi_w)-K\sqrt{\frac{\mathbb{E}_{x\sim\mathcal D}[D_x]}{2}}
\\
J_K(\pi_{\theta^+})
\ge
J_K(\pi_w)-K\sqrt{\frac{\varepsilon_{\mathrm{succ}}}{2}},
\end{gather}
where the final inequality uses Equation~\eqref{eq:rq1-binary-transfer}.
Combining this bound with the margin condition in Equation~\eqref{eq:rq1-positive-transfer-margin} proves the chained conclusion in Equation~\eqref{eq:rq1-center-transfer-bound}.
\FloatBarrier
\clearpage
\section{Complete Majority-Vote Results}
\label{app:majority-vote-results}

This appendix reports the Maj@16 and Maj@32 results corresponding to the
Pass@\(K\) results in Tables~\ref{tab:rq1-instruct-final},
\ref{tab:rq1-deepseek-math}, and~\ref{tab:rq2-deepseek-nonmath}.
Maj@\(K\) applies deterministic plurality voting after task-specific answer
normalization, as defined in Appendix~\ref{app:evaluation-metrics}.

\input{appendix_majority_tables}

\FloatBarrier
\clearpage
\section{Detailed Held-Out Performance Changes}
\label{app:heldout-performance-changes}

Table~\ref{tab:rq2-heldout-change} reports the metric-wise Easy Setting results underlying the held-out summary in Figure~\ref{fig:rq123-overview}(b).
For each model and metric, we compute post-training minus initial performance on each held-out benchmark and then average these changes equally over the five non-GSM8K tasks.

\begin{table}[h]
    \centering
    {\fontsize{8}{10}\selectfont
    
    \setlength{\tabcolsep}{6pt}
    \global\softtabdatafalse
    \begin{tabular}{@{}l*{5}{N}@{}}
    \softtabtoprule
    & \multicolumn{3}{c}{\textbf{Pass changes}} &
    \multicolumn{2}{c}{\textbf{Majority-vote changes}} \\
    \cmidrule(l{1pt}r{1pt}){2-4}\cmidrule(l{1pt}r{1pt}){5-6}
    Method & $\Delta$@1 & $\Delta$@16 & $\Delta$@32 & $\Delta$M@16 & $\Delta$M@32 \\
    \softtabheadrule
    \multicolumn{6}{l}{\textbf{Qwen2.5-1.5B-Instruct}} \\
    \global\softtabdatatrue \grpomethod & $+0.7$ & $-0.2$ & $-0.1$ & $+1.3$ & $+1.5$ \\
    \esmethod & $+0.0$ & $+0.6$ & $+0.8$ & $+1.0$ & $+2.0$ \\
    \midrule
    \multicolumn{6}{l}{\textbf{Llama-3.2-3B-Instruct}} \\
    \grpomethod & $+2.3$ & $-1.3$ & $-1.6$ & $+1.3$ & $+0.6$ \\
    \esmethod & $+1.4$ & $+2.2$ & $+2.2$ & $-0.0$ & $-0.0$ \\
    \midrule
    \multicolumn{6}{l}{\textbf{Qwen2.5-7B-Instruct}} \\
    \grpomethod & $+0.9$ & $-1.4$ & $-1.6$ & $-0.5$ & $-0.4$ \\
    \esmethod & $-0.5$ & $+0.3$ & $+0.2$ & $+0.3$ & $+0.7$ \\
    \softtabbottomrule
    \end{tabular}
    \global\softtabdatafalse
    \par}
    \caption{Held-out performance changes for matched Easy Setting runs, averaged equally over five non-GSM8K tasks and reported in percentage points. \textbf{M@16} and \textbf{M@32} denote Maj@16 and Maj@32. Positive values denote gains.}
    \label{tab:rq2-heldout-change}
\end{table}

Averaged across the three models, all five ES metric changes are positive, whereas GRPO decreases Pass@16 and Pass@32.

\FloatBarrier
\clearpage
\section{Complete Results for Magnitude-Thresholded ES}
\label{app:rq2-magnitude-threshold-results}

This appendix reports the complete target-task evaluation results for
magnitude-thresholded ES across all four models. For a threshold $\tau$,
coordinates satisfying
$0<|\Delta\theta_i|\leq\tau$ are set to zero, equivalently resetting the corresponding parameters to their Base Model values, and
update sparsity is computed over nonzero ES changes according to
Equation~\eqref{eq:rq2-update-sparsity}. Coordinates with zero change are
excluded from both the sparsity denominator and the ablated set. Pass@1 is the empirical single-sample accuracy
estimated from 32 retained responses per problem. The $\Delta$ Full ES column
reports the signed absolute difference from the corresponding unablated ES
endpoint on the same percentage scale.

\begin{table}[H]
    \centering
    \caption{Complete GSM8K Pass@1 results for magnitude-thresholded ES in the Easy Setting. Base Model values are included as absolute references.}
    \label{tab:app-rq2-magnitude-threshold-easy}
    \vspace{2pt}
    {\fontsize{8}{9.5}\selectfont
    \renewcommand{\arraystretch}{1.05}
    \setlength{\tabcolsep}{8pt}
    \begin{tabular}{@{}cccc@{}}
    \toprule
    Threshold $\tau$ & Update sparsity (\%) & Pass@1 (\%) & $\Delta$ Full ES (\%) \\
    \midrule
    \multicolumn{4}{l}{\textbf{Qwen2.5-1.5B-Instruct}} \\
    Full ES & 0.00 & 73.351 & $+0.000$ \\
    $2.5\times10^{-4}$ & 27.82 & 73.330 & $-0.021$ \\
    $5.0\times10^{-4}$ & 50.77 & 73.102 & $-0.249$ \\
    $1.0\times10^{-3}$ & 79.11 & 73.000 & $-0.351$ \\
    $1.5\times10^{-3}$ & 92.47 & 72.091 & $-1.260$ \\
    $2.0\times10^{-3}$ & 97.57 & 70.681 & $-2.670$ \\
    Base Model & 100.00 & 70.271 & -- \\
    \midrule
    \multicolumn{4}{l}{\textbf{Llama-3.2-3B-Instruct}} \\
    Full ES & 0.00 & 81.691 & $+0.000$ \\
    $2.5\times10^{-4}$ & 25.27 & 81.658 & $-0.033$ \\
    $5.0\times10^{-4}$ & 47.74 & 81.440 & $-0.251$ \\
    $1.0\times10^{-3}$ & 78.08 & 81.560 & $-0.130$ \\
    $1.5\times10^{-3}$ & 92.64 & 78.433 & $-3.258$ \\
    $2.0\times10^{-3}$ & 97.89 & 77.883 & $-3.807$ \\
    Base Model & 100.00 & 77.670 & -- \\
    \midrule
    \multicolumn{4}{l}{\textbf{Qwen2.5-7B-Instruct}} \\
    Full ES & 0.00 & 91.016 & $+0.000$ \\
    $2.5\times10^{-4}$ & 25.43 & 91.087 & $+0.071$ \\
    $5.0\times10^{-4}$ & 47.51 & 91.028 & $+0.012$ \\
    $1.0\times10^{-3}$ & 78.41 & 91.504 & $+0.488$ \\
    $1.5\times10^{-3}$ & 93.02 & 91.736 & $+0.720$ \\
    $2.0\times10^{-3}$ & 98.11 & 91.824 & $+0.808$ \\
    Base Model & 100.00 & 91.774 & -- \\
    \bottomrule
    \end{tabular}
    \par}
\end{table}

\begin{table}[H]
    \centering
    \caption{Complete MATH-500 Pass@1 results for magnitude-thresholded ES with DeepSeek-R1-Distill-Qwen-1.5B in the Hard Setting. The Base Model value is included as an absolute reference.}
    \label{tab:app-rq2-magnitude-threshold-hard}
    \vspace{2pt}
    {\fontsize{8}{9.5}\selectfont
    \renewcommand{\arraystretch}{1.05}
    \setlength{\tabcolsep}{8pt}
    \begin{tabular}{@{}cccc@{}}
    \toprule
    Threshold $\tau$ & Update sparsity (\%) & Pass@1 (\%) & $\Delta$ Full ES (\%) \\
    \midrule
    Full ES & 0.00 & 82.844 & $+0.000$ \\
    $2.5\times10^{-4}$ & 17.80 & 83.112 & $+0.269$ \\
    $5.0\times10^{-4}$ & 38.32 & 82.919 & $+0.075$ \\
    $1.0\times10^{-3}$ & 62.18 & 83.075 & $+0.231$ \\
    $1.5\times10^{-3}$ & 77.61 & 83.013 & $+0.169$ \\
    $2.0\times10^{-3}$ & 87.29 & 82.394 & $-0.450$ \\
    $2.5\times10^{-3}$ & 93.16 & 82.213 & $-0.631$ \\
    $3.0\times10^{-3}$ & 96.64 & 81.388 & $-1.456$ \\
    Base Model & 100.00 & 80.856 & -- \\
    \bottomrule
    \end{tabular}
    \par}
\end{table}

At comparable sparsity near 78\%, all four endpoints remain close to Full ES:
the Pass@1 changes are $-0.351$, $-0.130$, $+0.488$, and $+0.169$
percentage points for Qwen2.5-1.5B-Instruct, Llama-3.2-3B-Instruct,
Qwen2.5-7B-Instruct, and DeepSeek-R1-Distill-Qwen-1.5B, respectively.
Their behavior diverges as the threshold expands farther into the update
distribution. Qwen2.5-1.5B-Instruct and Llama-3.2-3B-Instruct degrade at
approximately 92\% sparsity, DeepSeek degrades more gradually, and Qwen2.5-7B
improves across the tested range.

\FloatBarrier
\section{ES--ZO Estimator Equivalence and Its Finite-Sample Boundary}
\label{app:estimator-boundary}

This appendix separates an ideal estimator identity from a finite-sample property of stochastic reasoning evaluations.
The identity shows that ES and ZO can estimate the same smoothed objective; the diagnostic tests whether paired evaluations retain enough covariance to reduce variance.
The Gaussian smoothing and score-function construction below are standard tools in gradient-free optimization and ES~\citep{nesterov2017random,salimans2017es}.

\subsection{A Unified Smoothed-Objective View}

For a measurable scalar objective \(F\), we use a maximization convention throughout.
For reward maximization, we take \(F(\theta)=R(\theta)\); for loss minimization, we take
\(F(\theta)=-\mathcal{L}(\theta)\), so that all estimators below are written in ascent form.
Assume that some neighborhood \(U\) of \(\theta\) satisfies
\(\sup_{\theta'\in U}\mathbb{E}_{\epsilon}[|F(\theta'+\sigma\epsilon)|(1+\lVert\epsilon\rVert)]<\infty\),
and define its Gaussian smoothing as
\begin{equation}
F_\sigma(\theta)
=
\mathbb{E}_{\epsilon\sim\mathcal{N}(0,I)}
\left[F(\theta+\sigma\epsilon)\right].
\label{eq:app-estimator-smoothed-objective}
\end{equation}
The Gaussian score-function identity gives
\begin{equation}
\nabla F_\sigma(\theta)
=
\frac{1}{\sigma}
\mathbb{E}_{\epsilon}
\left[F(\theta+\sigma\epsilon)\epsilon\right].
\label{eq:app-estimator-score-identity}
\end{equation}
The resulting one-point estimator over \(N\) directions is
\begin{equation}
\widehat{g}^{1p}
=
\frac{1}{N\sigma}
\sum_{i=1}^{N}
F(\theta+\sigma\epsilon_i)\epsilon_i,
\qquad
\mathbb{E}\!\left[\widehat{g}^{1p}\right]
=
\nabla F_\sigma(\theta).
\label{eq:app-one-point-estimator}
\end{equation}

The multi-direction two-point estimator is
\begin{equation}
\widehat{g}^{2p}
=
\frac{1}{2N\sigma}
\sum_{i=1}^{N}
\left[
F(\theta+\sigma\epsilon_i)
-
F(\theta-\sigma\epsilon_i)
\right]\epsilon_i.
\label{eq:app-two-point-estimator}
\end{equation}
Because \(\epsilon\) and \(-\epsilon\) have the same distribution,
\begin{equation}
\mathbb{E}_{\epsilon}
\left[F(\theta-\sigma\epsilon)\epsilon\right]
=
-
\mathbb{E}_{\epsilon}
\left[F(\theta+\sigma\epsilon)\epsilon\right].
\end{equation}
Substitution into Equation~\eqref{eq:app-two-point-estimator} yields
\begin{equation}
\mathbb{E}\!\left[\widehat{g}^{2p}\right]
=
\nabla F_\sigma(\theta)
=
\mathbb{E}\!\left[\widehat{g}^{1p}\right].
\label{eq:app-estimator-expectation-equivalence}
\end{equation}
If ES uses the same positive--negative perturbation pairs, its antithetic estimator is exactly Equation~\eqref{eq:app-two-point-estimator}.
Thus, antithetic ES and two-point ZO are the same estimator when they share the objective, perturbation distribution, scale, and raw paired-difference rule.
This identity does not imply equal finite-budget variance because one pair consumes two function evaluations, whereas one one-point sample consumes one.
Thus, \(N\) counts perturbation directions in both estimators, while their perturbed-evaluation counts are \(N_{\mathrm{eval}}=N\) and \(N_{\mathrm{eval}}=2N\), respectively.

\begin{table*}[!t]
    \centering
    {\fontsize{8}{9.5}\selectfont
    
    \setlength{\tabcolsep}{5pt}
    \textbf{(a) Objective comparison at \(\sigma=10^{-3}\)}\par
    \smallskip
    \begin{tabular}{llrrrr}
        \toprule
        Objective & Evaluation coupling &
        \(\operatorname{Corr}(X_+,X_0)\) &
        \(\kappa_{\mathrm{base}}\) &
        \(\operatorname{Corr}(X_+,X_-)\) &
        \(\kappa_{\mathrm{pair}}\) \\
        \midrule
        GSM8K regenerated-rollout reward & Common seed      & 0.2777 & 1.4445 & 0.1304 & 1.9870 \\
        GSM8K regenerated-rollout reward & Independent seed & 0.0561 & 1.3577 & 0.0224 & 2.7847 \\
        SST-2 supervised CE & Fixed batch     & 0.9776 & 0.0447 & 0.9196 & 0.1557 \\
        \midrule
        Mean token log-probability & Fixed rollout & 0.9939 & 0.0125 & 0.9857 & 0.0285 \\
        Policy sequence loss       & Fixed rollout & 0.9939 & 0.0167 & 0.9903 & 0.0194 \\
        GRPO surrogate             & Fixed rollout & 0.9998 & 0.0006 & 0.9997 & 0.0007 \\
        \bottomrule
    \end{tabular}

    \vspace{0.8em}
    \textbf{(b) Perturbation-scale sweep for GSM8K regenerated-rollout reward}\par
    \smallskip
    \setlength{\tabcolsep}{7pt}
    \begin{tabular}{llrrrr}
        \toprule
        \(\sigma\) & Seed coupling &
        \(\operatorname{Corr}(X_+,X_0)\) &
        \(\kappa_{\mathrm{base}}\) &
        \(\operatorname{Corr}(X_+,X_-)\) &
        \(\kappa_{\mathrm{pair}}\) \\
        \midrule
        \(10^{-4}\)          & Common      &  0.8797 & 0.2288 & 0.5537 & 0.8194 \\
        \(10^{-4}\)          & Independent & -0.0245 & 1.4080 & 0.0450 & 2.3822 \\
        \(3\times10^{-4}\)   & Common      &  0.5607 & 1.0424 & 0.3026 & 1.9434 \\
        \(3\times10^{-4}\)   & Independent & -0.0303 & 1.6687 & 0.0532 & 2.4379 \\
        \(10^{-3}\)          & Common      &  0.2777 & 1.4445 & 0.1304 & 1.9870 \\
        \(10^{-3}\)          & Independent &  0.0561 & 1.3577 & 0.0224 & 2.7847 \\
        \bottomrule
    \end{tabular}
    \par}
    \caption{Local subtraction diagnostic at the Qwen2.5-1.5B-Instruct base checkpoint. \(\kappa_{\mathrm{base}}\) and \(\kappa_{\mathrm{pair}}\) are the raw scalar variance ratios in Equation~\eqref{eq:app-raw-variance-ratios}; lower is better and values below \(1\) indicate variance reduction. Panel (a) compares objectives at \(\sigma=10^{-3}\), and Panel (b) sweeps the scale for GSM8K regenerated-rollout reward.}
    \label{tab:app-estimator-diagnostic}
\end{table*}

\subsection{Covariance Required by Paired Subtraction}

Suppose an objective evaluation is stochastic because it includes prompt sampling, autoregressive generation, or verifier noise.
Let \(\xi\) collect this evaluation randomness and let \(\widehat F(\vartheta;\xi)\) be an unbiased scalar evaluation of the objective, such that
\begin{equation}
\mathbb{E}_{\xi}\!\left[\widehat F(\vartheta;\xi)\right]=F(\vartheta)
\label{eq:app-unbiased-stochastic-evaluation}
\end{equation}
for every evaluated parameter point \(\vartheta\).
For fixed \((\theta,\epsilon,\sigma)\), the repeated evaluations are
\begin{equation}
\begin{gathered}
X_+=\widehat F(\theta+\sigma\epsilon;\xi_+),\\
X_0=\widehat F(\theta;\xi_0),\\
X_-=\widehat F(\theta-\sigma\epsilon;\xi_-).
\end{gathered}
\label{eq:app-evaluation-definition}
\end{equation}
The random variables \((\xi_+,\xi_0,\xi_-)\) may be coupled through common random numbers or sampled independently.
Their raw difference variances satisfy
\begin{gather}
\operatorname{Var}(X_+-X_0)
=
\operatorname{Var}(X_+)+\operatorname{Var}(X_0)
-2\operatorname{Cov}(X_+,X_0),
\label{eq:app-baseline-variance}
\\
\operatorname{Var}(X_+-X_-)
=
\operatorname{Var}(X_+)+\operatorname{Var}(X_-)
-2\operatorname{Cov}(X_+,X_-).
\label{eq:app-two-point-variance}
\end{gather}
In particular, \(\operatorname{Var}(X_+-X_-)<\operatorname{Var}(X_+)\) if and only if
\begin{equation}
\operatorname{Cov}(X_+,X_-)>\frac{1}{2}\operatorname{Var}(X_-).
\label{eq:app-two-point-covariance-condition}
\end{equation}
We report the correlations and the raw variance ratios
\begin{equation}
\kappa_{\mathrm{base}}
=
\frac{\operatorname{Var}(X_+-X_0)}{\operatorname{Var}(X_+)},
\qquad
\kappa_{\mathrm{pair}}
=
\frac{\operatorname{Var}(X_+-X_-)}{\operatorname{Var}(X_+)}.
\label{eq:app-raw-variance-ratios}
\end{equation}
A ratio below one means that subtraction reduces the variance of the scalar objective values in this diagnostic.

\subsection{Local Perturbation Diagnostic}
\label{app:estimator-diagnostic}

\paragraph{Protocol.}
We conduct a local diagnostic around the Qwen2.5-1.5B-Instruct base checkpoint.
For each sampled full-parameter direction, we evaluate Equation~\eqref{eq:app-evaluation-definition} at several perturbation scales.
The reported statistics pool scalar observations across prompts, generation seeds, and directions.

\paragraph{Objectives.}
We compare three evaluation protocols.
First, \emph{GSM8K regenerated-rollout reward} regenerates a response independently at each of the positive, center, and negative parameter points and assigns a terminal correctness reward.
Second, \emph{SST-2 supervised CE} evaluates a differentiable cross-entropy objective on the same fixed batch at all three parameter points.
Third, \emph{GSM8K fixed rollout} generates responses once at the base checkpoint, freezes them, and recomputes mean token log-probability, policy sequence loss, and a GRPO surrogate at the perturbed points.
Fixed rollouts isolate rescoring from autoregressive generation noise.

\paragraph{Randomness Coupling.}
For regenerated rollouts, \emph{common} uses the same generation seed at the three parameter points, whereas \emph{independent} uses different seeds.
Common seeds implement common random numbers, but they do not guarantee identical autoregressive trajectories after a parameter perturbation changes an early token distribution.

\subsection{Diagnostic Results}

Table~\ref{tab:app-estimator-diagnostic}(a) compares the three objective classes at \(\sigma=10^{-3}\).
The repeated-direction regenerated-rollout run uses \(16\) prompts, \(16\) generation seeds per prompt, and \(4\) perturbation directions.
The fixed-rollout run uses \(16\) prompts, \(8\) seeds per prompt, and \(4\) directions.
The supervised control contains \(128\) pooled scalar observations.

GSM8K regenerated-rollout rewards are weakly correlated across the positive and negative perturbations at this scale, and paired subtraction increases their raw scalar variance.
The same perturbation implementation yields strong correlations and ratios below one for supervised CE and all fixed-rollout objectives.
This contrast localizes the observed instability to the combination of regenerating autoregressive responses and evaluating terminal rewards rather than to paired parameter perturbations alone.
The very small fixed-rollout GRPO ratio further shows that a GRPO surrogate is not intrinsically incompatible with paired ZO evaluation when the responses are held fixed.

Table~\ref{tab:app-estimator-diagnostic}(b) shows that coupling regenerated-rollout reward evaluations with common random numbers is scale dependent.
At \(\sigma=10^{-4}\), common seeds produce \(\kappa_{\mathrm{pair}}=0.8194\), whereas independent seeds remain above one.
At the two larger scales, paired subtraction does not reduce raw variance under either seed mode.

\FloatBarrier

\subsection{Scope of the Evidence}

The diagnostic supports a conclusion: near the tested base checkpoint, regenerated-rollout correctness rewards provide substantially weaker positive--negative coupling than fixed-batch supervised or fixed-rollout objectives, so raw paired subtraction does not reliably cancel evaluation variance except at small scales.

Figure~\ref{fig:rq123-overview}(c), item~(3), complements the local diagnostic with a three-epoch comparison under matched perturbed evaluations and reward normalization.
\section{Population-Size Scaling Trajectories}
\label{app:population-scaling-trajectories}

\begin{figure*}[t]
    \centering
    \includegraphics[width=0.42\textwidth]{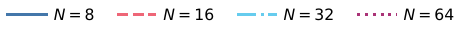}\par
    \vspace{1pt}
    \begin{subfigure}[t]{0.98\textwidth}
        \centering
        \includegraphics[width=\linewidth]{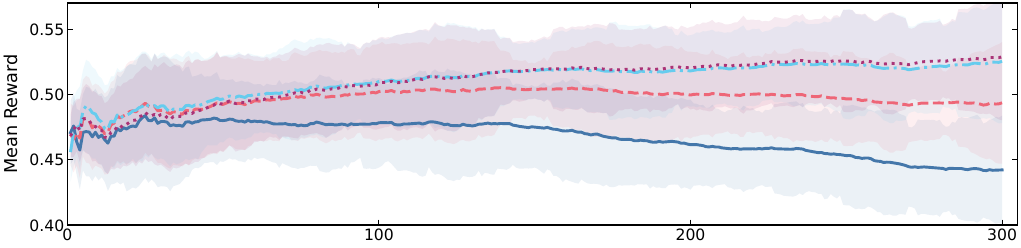}
        \caption{\textbf{Qwen2.5-0.5B-Instruct}}
        \label{fig:app-rq3-population-qwen05}
    \end{subfigure}\par
    \vspace{1pt}
    \begin{subfigure}[t]{0.98\textwidth}
        \centering
        \includegraphics[width=\linewidth]{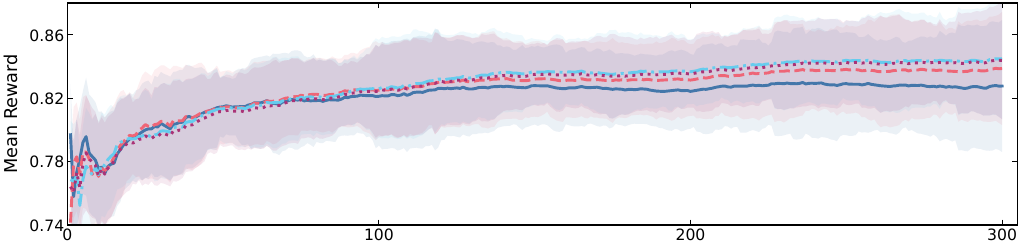}
        \caption{\textbf{Qwen2.5-1.5B-Instruct}}
        \label{fig:app-rq3-population-qwen15}
    \end{subfigure}\par
    \vspace{1pt}
    \begin{subfigure}[t]{0.98\textwidth}
        \centering
        \includegraphics[width=\linewidth]{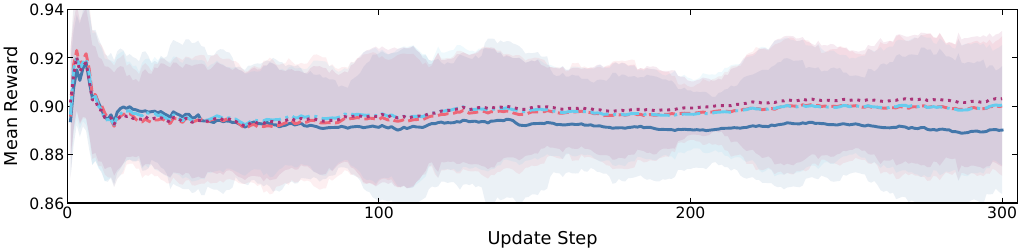}
        \caption{\textbf{Qwen2.5-3B-Instruct}}
        \label{fig:app-rq3-population-qwen3}
    \end{subfigure}
    \caption{Full GSM8K population-size trajectories across three model scales. Lines show debiased exponential smoothing with weight \(0.99\). Same-color shading shows the local standard deviation of raw-minus-smoothed residuals; it describes within-trajectory fluctuation. Model-specific vertical ranges expose within-model differences.}
    \label{fig:app-rq3-population-trajectories}
\end{figure*}

Figure~\ref{fig:app-rq3-population-trajectories} reports the complete per-update mean-reward trajectories summarized by Table~\ref{tab:rq3-population-checkpoints}.
The colored curves use debiased exponential smoothing with weight \(0.99\).
Distinct colors and line styles identify \(N\in\{8,16,32,64\}\).
The same-color shaded region spans one local standard deviation of the raw-minus-smoothed residuals.
This band visualizes within-trajectory fluctuation.

For Qwen2.5-0.5B-Instruct, stable late-stage improvement requires a larger population: \(N=8\) and \(N=16\) decline over later updates, whereas \(N=32\) and \(N=64\) continue improving.
For Qwen2.5-1.5B-Instruct and Qwen2.5-3B-Instruct, the \(N=16\), \(N=32\), and \(N=64\) curves remain substantially closer throughout training.

\FloatBarrier
\section{Experimental Details}
\label{app:experimental-details}

This appendix records the settings used to produce the reported results.
We separate training hyperparameters, evaluation settings, prompt templates, and dataset provenance.
The locally trained ES/GRPO runs use full-parameter updates.
The training and testing hardware used NVIDIA A100-SXM4-80GB GPUs.

\subsection{Implementation-Level Training Procedure}
\label{app:implementation-procedures}

\paragraph{Full-parameter one-point ES.}
The following procedure describes the implementation used for all locally trained ES models.
It is the practical one-point, reward-standardized update studied in the paper.
The Easy and Hard settings use the same update procedure and differ only in the model, dataset, epoch budget, and generation settings listed in Table~\ref{tab:app-training-hyperparameters}.

\begin{tcolorbox}[
  enhanced,
  breakable,
  colframe=black!55,
  colback=black!2,
  boxrule=0.5pt,
  arc=1mm,
  left=4pt,
  right=4pt,
  top=3pt,
  bottom=3pt,
  fontupper=\small,
  before skip=5pt,
  after skip=5pt
]
\textbf{Implementation Procedure: Full-parameter one-point ES training}
\begin{enumerate}[leftmargin=*,itemsep=1pt,topsep=2pt]
    \item Initialize the center model \(\theta_0\), shuffle the training split, and construct one keep-tail mini-batch schedule per epoch.
    \item At update \(t\), draw \(N\) independent 32-bit perturbation seeds. A seed and the stable name of each parameter tensor deterministically reconstruct an independent standard-Gaussian tensor, yielding a full-model direction \(\epsilon_i\).
    \item Distribute the \(N\) directions over eight single-GPU vLLM engines. For direction \(i\), add \(\sigma\epsilon_i\) to every model parameter in place, generate one response for every prompt in the shared mini-batch, compute the task-verifier rewards, average them into \(R_i\), and then subtract the same seeded perturbation to restore the center.
    \item Standardize the population rewards with
    \[
    z_i=\frac{R_i-\bar R}{\sqrt{N^{-1}\sum_{j=1}^{N}(R_j-\bar R)^2}+\varepsilon_{\mathrm{num}}}.
    \]
    \item Reconstruct the same directions and update every engine locally by
    \[
    \theta_{t+1}
    =
    \theta_t+\frac{\alpha}{N}\sum_{i=1}^{N}z_i\epsilon_i.
    \]
    Here \(\alpha\) is the center-update scale in Table~\ref{tab:app-training-hyperparameters}.
\end{enumerate}
\end{tcolorbox}

The response-level ES training reward is computed by the task-specific verifier.
For the reported mathematical runs, a verifier-correct response receives reward \(1\); a response that follows the required final-answer format but is incorrect receives the configured format reward \(0.1\); and an answer that fails the required format receives \(0\).

\subsection{ES--GRPO FLOP Matching}
\label{app:es-grpo-flop-matching}

We match the estimated model FLOPs of ES and GRPO following the standard accounting used by \citet{gan2026neuralthickets}.
For a model with \(P\) parameters and a response length of \(L\) tokens, one forward pass requires approximately \(2PL\) FLOPs and one backward pass requires approximately \(4PL\) FLOPs.
A GRPO response requires a policy forward, a reference-policy forward, and a policy backward, giving
\begin{equation}
\operatorname{FLOPs}_{\mathrm{GRPO}}
=
8T_{\mathrm{GRPO}}B G P L,
\label{eq:app-grpo-flops}
\end{equation}
where \(T_{\mathrm{GRPO}}\) is the number of updates, \(B\) is the prompt batch size, and \(G\) is the number of responses per prompt.
ES evaluates each population member using only a forward pass, giving
\begin{equation}
\operatorname{FLOPs}_{\mathrm{ES}}
=
2T_{\mathrm{ES}}N D P L,
\label{eq:app-es-flops}
\end{equation}
where \(N\) is the population size and \(D\) is the number of prompts evaluated per direction.
With matched update counts \(T_{\mathrm{ES}}=T_{\mathrm{GRPO}}\), matched prompt batches \(D=B\), and \(N=32=4G\) for \(G=8\), Equations~\eqref{eq:app-grpo-flops} and~\eqref{eq:app-es-flops} are equal.
Thus, the reported ES and GRPO configurations are FLOP-matched under this model-compute accounting.

\subsection{Training Hyperparameters}
\label{app:training-hyperparameters}

Table~\ref{tab:app-training-hyperparameters} summarizes the training configurations used for the Easy and Hard ES/GRPO comparisons.

\begin{table}[H]
    \centering
    {\fontsize{8}{9.5}\selectfont
    
    \setlength{\tabcolsep}{2.2pt}
    \begin{tabularx}{\textwidth}{
        >{\raggedright\arraybackslash}p{0.16\textwidth}
        >{\centering\arraybackslash}X
        >{\centering\arraybackslash}X
        >{\centering\arraybackslash}X
        >{\centering\arraybackslash}X}
    \toprule
    Parameter & Easy ES & Easy GRPO & Hard ES & Hard GRPO \\
    \midrule
    Model &
    \makecell{Qwen2.5-1.5B-Instruct,\\Llama-3.2-3B-Instruct,\\Qwen2.5-7B-Instruct} &
    Same as Easy ES &
    \makecell{DeepSeek-R1-Distill-\\Qwen-1.5B} &
    Same as Hard ES \\
    Training data &
    GSM8K &
    GSM8K &
    DeepScaleR &
    DeepScaleR \\
    Epochs &
    2 &
    2 &
    1 &
    1 \\
    Global train batch &
    64 &
    64 &
    64 &
    64 \\
    Population / rollout group &
    32 directions &
    8 responses &
    32 directions &
    8 responses \\
    Learning scale &
    \(\sigma=1.5{\times}10^{-3}\), \(\alpha=2.5{\times}10^{-4}\) &
    \(1.0{\times}10^{-6}\) learning rate &
    \(\sigma=1.5{\times}10^{-3}\), \(\alpha=2.5{\times}10^{-4}\) &
    \(1.0{\times}10^{-6}\) learning rate \\
    Train-time sampling &
    \(\tau=0\) &
    \(\tau=1.0\) &
    \(\tau=0.6\) &
    \(\tau=1.0\) \\
    Response limit &
    2,048 &
    2,048 &
    8,192 &
    8,192 \\
    Reward normalization &
    Population \(z\)-score &
    Group-relative advantages &
    Population \(z\)-score &
    Group-relative standardized advantages \\
    GRPO update details &
    -- &
    \makecell{PPO minibatch 32; microbatch 2;\\clip 0.2; KL coefficient 0.001} &
    -- &
    \makecell{clip 0.2; KL coefficient 0.001;\\verl with vLLM rollout} \\
    \bottomrule
    \end{tabularx}
    \par}
    \caption{Training settings used for the reported comparisons. Hard GRPO is the released discrete-token GRPO checkpoint. Its training parameters are transcribed from Appendix~D.1 of the SofT-GRPO paper~\citep{zheng2026softgrposurpassingdiscretetokenllm}.}
    \label{tab:app-training-hyperparameters}
\end{table}

\subsection{Evaluation Configuration}
\label{app:evaluation-configuration}

\begin{wraptable}{r}{0.48\columnwidth}
    \centering
    {\fontsize{8}{9.5}\selectfont
    \setlength{\tabcolsep}{4pt}
    
    \begin{tabular}{@{}lcc@{}}
    \toprule
    Parameter & Easy & Hard \\
    \midrule
    Samples per problem & 32 & 32 \\
    Temperature & 0.6 & 0.6 \\
    Maximum new tokens & 2,048 & 8,192 \\
    Reported \(K\) & 16, 32 & 16, 32 \\
    Execution & Single GPU & Single GPU \\
    \bottomrule
    \end{tabular}
    \par}
    \caption{Offline sampled-evaluation settings.}
    \label{tab:app-evaluation-hyperparameters}
\end{wraptable}
Table~\ref{tab:app-evaluation-hyperparameters} summarizes the offline evaluation configurations used for the Easy and Hard comparisons.
Both protocols retain 32 responses per problem at temperature \(0.6\) and report \(K\in\{16,32\}\).
The Easy Setting limits each response to 2,048 new tokens, whereas the Hard Setting allows 8,192 to accommodate longer reasoning traces.
Both protocols run on a single GPU.
The Easy protocol additionally includes a one-sample greedy evaluation at temperature \(0\); the main Pass@\(K\) and majority-vote tables use the sampled protocol shown here.

\WFclear
\subsection{Prompt Templates}
\label{app:prompt-templates}

The following messages are rendered with each model's official chat template.

\begin{promptbox}{Prompt for Mathematical Reasoning}
\textcolor{promptrole}{system}\\
Please reason step by step, and put your final answer within
\texttt{\textbackslash boxed\{\}}.\\[2pt]
\textcolor{promptrole}{user}\\
\textcolor{promptvariable}{\{Question\}}
\end{promptbox}

\begin{promptbox}{Prompt for Multiple-Choice Reasoning}
\textcolor{promptrole}{system}\\
Please reason step by step to solve the following multiple-choice question.
Put your final choice letter inside \texttt{\textbackslash boxed\{\}}, e.g.,
\texttt{\textbackslash boxed\{C\}}.\\[2pt]
\textcolor{promptrole}{user}\\
\textcolor{promptvariable}{\{Question and labeled choices\}}
\end{promptbox}

\begin{promptbox}{Prompt for Open Question Answering}
\textcolor{promptrole}{system}\\
Answer the question using the provided context.
Keep the final answer short, and put it inside
\texttt{<answer> </answer>} tags.\\[2pt]
\textcolor{promptrole}{user}\\
\textcolor{promptvariable}{\{Context and question\}}
\end{promptbox}

\begin{promptbox}{Prompt for MBPP Code Generation}
\textcolor{promptrole}{system}\\
You are a Python programming assistant.
Write clean, correct Python code that passes the given tests.
The function name and signature must match exactly as specified in the test cases.
The final code block must not include the tests; do not include the tests in your answer.
Do not include explanations or text outside the code block.
Output exactly one Python markdown code block containing only the final solution code.\\[2pt]
\textcolor{promptrole}{user}\\
Task: \textcolor{promptvariable}{\{Programming task\}}\\
Test Cases: \textcolor{promptvariable}{\{Public tests\}}
\end{promptbox}

Countdown uses the role/content messages distributed with the dataset and renders them without rewriting the task statement.
Answer extraction follows the same task-family contract as scoring: boxed answers for mathematics and multiple choice, \texttt{<answer>} tags for open QA and Countdown, and one Python code block for MBPP.

\subsection{Datasets and Setting Rationale}
\label{app:dataset-provenance}

\begin{table*}[t]
    \centering
    {\fontsize{8}{9.5}\selectfont
    
    \setlength{\tabcolsep}{3pt}
    \begin{tabularx}{\textwidth}{
        >{\centering\arraybackslash}p{0.09\textwidth}
        >{\raggedright\arraybackslash}p{0.12\textwidth}
        >{\raggedright\arraybackslash}p{0.24\textwidth}
        >{\raggedright\arraybackslash}X}
    \toprule
    Setting & Role & Dataset / split & Purpose \\
    \midrule
    \multirow[c]{3}{*}[-18pt]{Easy} & Post-training & GSM8K train &
    Standard grade-school mathematical word problems with rule-verifiable numeric answers. \\
    & In-domain test & GSM8K test &
    Measures transfer to unseen problems from the post-training task family. \\
    & Held-out test &
    CommonsenseQA validation; HotpotQA distractor validation; Countdown test; GPQA Diamond; MBPP sanitized test &
    Covers commonsense choice, multi-hop QA, arithmetic planning, scientific reasoning, and code generation. \\
    \midrule
    \multirow[c]{3}{*}[-14pt]{Hard} & Post-training & DeepScaleR Preview &
    Mathematical RL training data used in the DeepScaleR line of work. \\
    & Mathematical test &
    AIME 2024; AIME 2025; AMC 2023; MATH-500 &
    Competition and advanced mathematical reasoning benchmarks commonly used for 1.5B reasoning-RL evaluation. \\
    & Held-out test &
    GPQA Diamond; MBPP sanitized test; CommonsenseQA validation; Countdown test &
    Tests whether mathematical post-training transfers to or degrades non-training task families. \\
    \bottomrule
    \end{tabularx}
    \par}
    \caption{Datasets and splits used in the reported comparisons. No held-out dataset contributes post-training rewards.}
    \label{tab:app-datasets}
\end{table*}

\begin{table*}[t]
    \centering
    {\fontsize{8}{9.5}\selectfont
    
    \setlength{\tabcolsep}{4pt}
    \begin{tabularx}{\textwidth}{
        >{\raggedright\arraybackslash}p{0.16\textwidth}
        >{\raggedright\arraybackslash}p{0.39\textwidth}
        >{\raggedright\arraybackslash}X}
    \toprule
    Dataset & Distributed source used for provenance & License declared by the source \\
    \midrule
    GSM8K &
    \texttt{openai/gsm8k} &
    MIT \\
    CommonsenseQA &
    \texttt{tau/commonsense\_qa} &
    MIT \\
    HotpotQA &
    Official HotpotQA dataset release &
    CC BY-SA 4.0 \\
    Countdown &
    \texttt{HuggingFaceTB/Countdown-Task-GOLD}; the fixed test construction corresponds to \texttt{Jiayi-Pan/Countdown-Tasks-3to4} &
    No license specified in either dataset card \\
    GPQA &
    \texttt{Idavidrein/gpqa} &
    CC BY 4.0 \\
    MBPP &
    \texttt{google-research-datasets/mbpp} &
    CC BY 4.0 \\
    DeepScaleR Preview &
    \texttt{agentica-org/DeepScaleR-Preview-Dataset} &
    MIT \\
    AIME 2024 / AIME 2025 &
    \texttt{math-ai/aime24}; \texttt{math-ai/aime25} &
    Apache-2.0 \\
    AMC 2023 &
    \texttt{math-ai/amc23} &
    No license specified in the dataset card \\
    MATH-500 &
    \texttt{HuggingFaceH4/MATH-500} &
    No license specified in the dataset card \\
    \bottomrule
    \end{tabularx}
    \par}
    \caption{Licenses declared by the dataset distributions used in this work.}
    \label{tab:app-dataset-licenses}
\end{table*}

The Easy Setting uses GSM8K because it is a widely used, readily verifiable mathematical-reasoning task and supports a controlled comparison across three instruction-tuned model scales~\citep{cobbe2021gsm8k}.
Its held-out suite uses CommonsenseQA~\citep{talmor2019commonsenseqa}, HotpotQA~\citep{yang2018hotpotqa}, Countdown-Task-GOLD~\citep{huggingfacetb2025countdown}, GPQA~\citep{rein2023gpqa}, and MBPP~\citep{austin2021mbpp} to test cross-domain retention.

The Hard Setting is a harder \emph{post-training protocol}, not a claim that every included item is uniformly harder than every GSM8K item.
It combines the DeepSeek-R1-Distill-Qwen-1.5B backbone with DeepScaleR mathematical RL training data~\citep{luo2025deepscaler}.
This choice aligns the training regime with DeepScaleR and the model/evaluation regime with representative mathematical GRPO work.
In particular, JustRL uses the same DeepSeek 1.5B backbone and reports AIME 2024, AIME 2025, AMC 2023, and MATH-500 among its standard mathematical benchmarks~\citep{he2025justrl}.
The Hard mathematical suite uses the math-ai AIME and AMC releases~\citep{mathai2025aime24,mathai2025aime25,mathai2025amc23} and MATH-500~\citep{lightman2023verify}.

\subsection{Evaluation Metrics}
\label{app:evaluation-metrics}

For a problem with \(M=32\) retained responses and \(c\) verifier-correct responses, Pass@1 is the average per-response correctness \(c/M\).
For \(K\le M\), Pass@\(K\) uses the standard without-replacement estimator
\begin{equation}
\widehat{\operatorname{Pass@}K}
=
1-\frac{\binom{M-c}{K}}{\binom{M}{K}},
\end{equation}
so Pass@32 is the fraction of problems with at least one correct response.
Maj@\(K\) normalizes task-specific answers, takes a deterministic plurality vote over the first \(K\) retained responses, treats failed extraction as abstention, and scores the selected answer against the gold answer.
Reported task scores are arithmetic means over problems, with each problem weighted equally; multi-task averages likewise assign equal weight to each benchmark.
\section{Related Work}
\label{app:related-work}

\subsection{Evolution Strategies for LLM Reasoning Post-Training}

Evolution strategies (ES) are population-based black-box optimizers that estimate parameter updates from forward-evaluated perturbations and naturally distribute candidate evaluations across workers~\citep{salimans2017es}.
Recent work has adapted this paradigm to LLM reasoning post-training.
Qiu et al.\ demonstrate that full-parameter ES can operate directly in billion-dimensional parameter spaces and optimize LLMs with outcome-level rewards~\citep{qiu2025esatscale}.
ESSA instead restricts evolutionary search to low-rank adapters and further reduces the search space by optimizing their singular values, enabling quantized forward-only alignment on mathematical reasoning and instruction-following tasks~\citep{korotyshova2025essa}.
EGGROLL structures perturbations as low-rank matrices to improve the arithmetic intensity of population evaluation and applies the resulting system to reasoning post-training at large population sizes~\citep{sarkar2025hyperscale}.
ESSAM combines full-parameter ES with a sharpness-aware objective and evaluates this design on GSM8K, emphasizing generalization and memory-efficient reasoning fine-tuning~\citep{sun2026essam}.
These methods establish several viable implementations of ES for LLM post-training, spanning direct full-parameter search, parameter-efficient search, and structured perturbations.

A related line of work asks whether the geometry of ES updates predicts capability retention.
Abdi et al.\ report that ES updates have larger norms and affect a larger fraction of parameter elements than GRPO updates~\citep{abdi2026catastrophic}.
They associate these update properties with held-out degradation.
However, their forgetting result is obtained from a restricted experimental configuration comprising a small training set, one model, one training task, and one held-out benchmark, making training-set overfitting difficult to distinguish from broad capability forgetting.
Hoy et al.\ provide complementary evidence that larger parameter movement does not necessarily lead to consistent behavioral degradation: even when ES and GRPO attain similar target-task accuracy, ES travels much farther in weight space and induces broader off-task KL shifts, while the resulting accuracy changes vary across held-out benchmarks and ES iteration budgets~\citep{hoy2026matching}.
Taken together, these studies show that large parameter displacement can accompany capability degradation, but neither weight-space distance nor policy KL has been established as a general predictor, much less a cause, of forgetting.

\subsection{Reinforcement Learning with Verifiable Rewards}

Reinforcement learning with verifiable rewards (RLVR) improves reasoning models using automatically checked outcome rewards from domains such as mathematics and code.
Group Relative Policy Optimization (GRPO) removes the learned critic by normalizing rewards within a group of responses and optimizing a clipped token-level policy objective~\citep{shao2024deepseekmath}.
Although this recipe can improve single-sample accuracy, recent work identifies policy-entropy collapse as a recurring limitation of RLVR.
The entropy-mechanism analysis relates entropy change under policy-gradient updates to a probability--update covariance and observes sustained entropy decrease during reasoning training~\citep{cui2025entropymechanism}.
Subsequent studies attribute collapse to factors including positive-advantage tokens, imbalanced token-level entropy flows, clipping and off-policy reuse, and premature confidence at a small set of structurally important decision points~\citep{xu2026entropyflow,petrenko2026entropy,jin2026revisiting,jang2026badapples}.

A second line of work evaluates coverage behaviorally through large-budget sampling and asks whether RLVR expands the set of correct reasoning paths accessible from the base model.
Large-budget sampling studies find that RLVR can improve Pass@1 while underperforming the base model at larger \(K\), indicating more efficient exploitation of high-probability correct paths but weaker coverage of less likely solutions~\citep{yue2025doesrlincentivize}.
Controlled pretraining experiments further show that RL post-training tends to amplify behaviors already represented in the pretraining distribution~\citep{zhao2025echochamber}.
Support-based analyses similarly find that RLVR may shrink access to some correct solutions even as it raises single-attempt accuracy~\citep{wu2025invisibleleash}.
These findings motivate evaluating Pass@1 together with Pass@\(K\), rather than interpreting an increase in single-sample accuracy as an unconditional expansion of reasoning capability.

\subsection{Zeroth-Order Optimization for LLMs}

Zeroth-order (ZO) optimization replaces backpropagated derivatives with a gradient-shaped estimate constructed solely from scalar objective values.
Given a Gaussian direction \(u\sim\mathcal{N}(0,I)\) and perturbation radius \(\mu\), two forward evaluations first estimate the directional derivative as \(\delta_u f(\theta)=[f(\theta+\mu u)-f(\theta-\mu u)]/(2\mu)\), then map this scalar back to parameter space as the pseudo-gradient \(\widehat{g}_{\mathrm{ZO}}(\theta;u)=\delta_u f(\theta)u\).
Averaging these pseudo-gradients across directions estimates the gradient of a Gaussian-smoothed objective, enabling parameter updates without differentiating through the model~\citep{nesterov2017random}.
MeZO adapts simultaneous perturbation-based ZO optimization to operate in place, reducing memory to approximately the inference footprint and supporting both full-parameter and parameter-efficient language-model fine-tuning~\citep{malladi2023mezo}.
Variance-reduced extensions such as MeZO-SVRG improve the stability and convergence of forward-only language-model fine-tuning by periodically constructing lower-variance update estimates~\citep{gautam2024variance}.
ZO-Bench broadens the comparison across model families, task types, and fine-tuning schemes, and studies blockwise descent, hybrid optimization, forward-gradient estimators, and gradient sparsity~\citep{zhang2024zobench}.

ES and ZO are not disjoint estimator families.
With the same objective, perturbation distribution, scale, and raw positive--negative difference, antithetic ES is algebraically identical to a two-point ZO estimator.
The practical distinction in LLM reasoning arises from the evaluation protocol: much of the ZO fine-tuning literature studies fixed downstream objectives, whereas reasoning post-training repeatedly generates stochastic autoregressive responses and scores them with terminal verifiers.
As analyzed in Appendix~\ref{app:estimator-boundary}, resampling responses at the perturbed parameter points can weaken the covariance required for paired subtraction to reduce variance.
This boundary makes one- versus two-point estimation an empirical design choice under rollout rewards, rather than one whose behavior can be inferred unchanged from fixed supervised objectives.

\clearpage
\section{Pass@K Profiles of Individual Post-Training Methods and Sequential Compositions}
\label{app:es-grpo-fusion-pass-at-k}

\begin{figure}[H]
    \centering
    \includegraphics[width=0.60\linewidth]{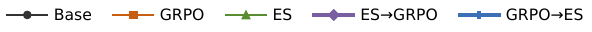}
    \vspace{0.15em}

    \begin{subfigure}[t]{0.49\linewidth}
        \centering
        \includegraphics[width=\linewidth]{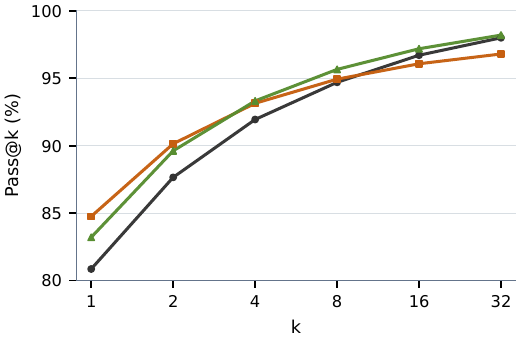}
        \caption{\textbf{MATH-500}\\[1pt]
        \subfigcontext{DeepSeek-R1-Distill-Qwen-1.5B}}
        \label{fig:fusion-math500}
    \end{subfigure}\hfill
    \begin{subfigure}[t]{0.49\linewidth}
        \centering
        \includegraphics[width=\linewidth]{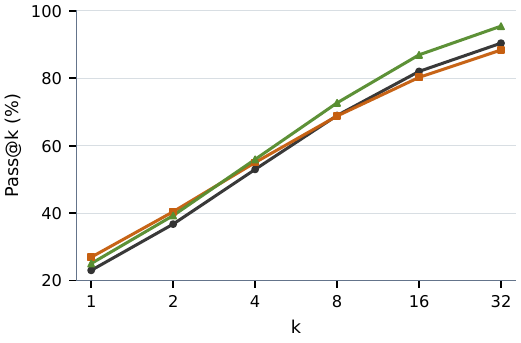}
        \caption{\textbf{GPQA}\\[1pt]
        \subfigcontext{Llama-3.2-3B-Instruct}}
        \label{fig:fusion-llama-gpqa}
    \end{subfigure}

    \vspace{0.35em}

    \begin{subfigure}[t]{0.49\linewidth}
        \centering
        \includegraphics[width=\linewidth]{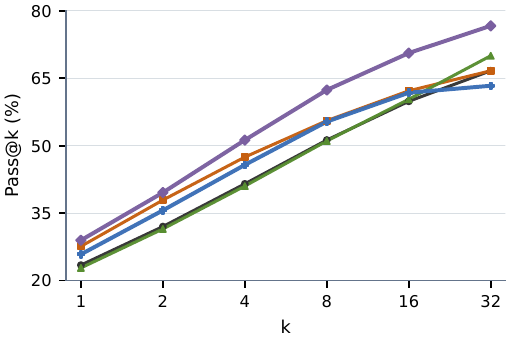}
        \caption{\textbf{AIME24}\\[1pt]
        \subfigcontext{DeepSeek-R1-Distill-Qwen-1.5B}}
        \label{fig:fusion-aime24}
    \end{subfigure}\hfill
    \begin{subfigure}[t]{0.49\linewidth}
        \centering
        \includegraphics[width=\linewidth]{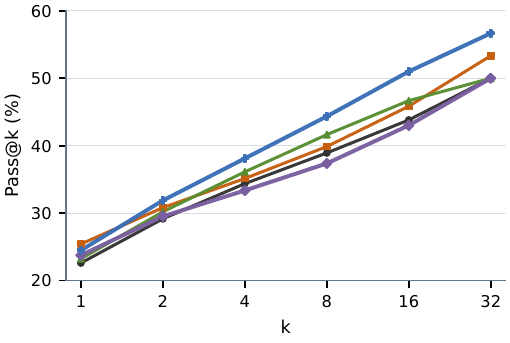}
        \caption{\textbf{AIME25}\\[1pt]
        \subfigcontext{DeepSeek-R1-Distill-Qwen-1.5B}}
        \label{fig:fusion-aime25}
    \end{subfigure}

    \caption{Pass@\(K\) profiles of individual post-training methods and sequential compositions. For individual post-training, GRPO leads at \(K=1\), whereas ES overtakes it at larger \(K\) while remaining above the Base Model throughout. Among the two sequential compositions, ES$\rightarrow$GRPO gives the strongest AIME24 profile, while GRPO$\rightarrow$ES leads on AIME25 for \(K\geq2\), showing complementary but order-dependent benefits.}
    \label{fig:fusion-selected-pass-at-k}
\end{figure}

Figure~\ref{fig:fusion-selected-pass-at-k} first compares the Pass@\(K\) profiles of ES and GRPO when applied individually, and then examines whether the two sequential compositions can exploit their complementary strengths.
In panels (a--b), GRPO leads at Pass@1, whereas ES overtakes it as \(K\) grows while remaining above the Base Model across all reported \(K\) values.
In panels (c--d), ES$\rightarrow$GRPO gives the strongest AIME24 profile, while GRPO$\rightarrow$ES leads on AIME25 for \(K\geq2\).
Thus, the two sequential compositions can better exploit the complementary strengths of ES and GRPO, although the effective training order is task-dependent.

\end{document}

%% file: figures/rq3_population_scaling_checkpoints.tex
% Generated by plot_rq3_population_scaling_mean_reward.py.
\newcommand{\rqPopulationCheckpointRows}{%
        100 & 0.4782 & \cellcolor{gray!25}{0.5021} & \cellcolor{gray!25}{0.5100} & 0.5085 & \cellcolor{gray!25}{0.8211} & \cellcolor{gray!25}{0.8250} & \cellcolor{gray!25}{0.8259} & 0.8244 & \cellcolor{gray!25}{0.8912} & \cellcolor{gray!25}{0.8945} & \cellcolor{gray!25}{0.8960} & 0.8957 \\
        200 & 0.4614 & 0.4995 & \cellcolor{gray!25}{0.5180} & 0.5201 & 0.8242 & \cellcolor{gray!25}{0.8314} & \cellcolor{gray!25}{0.8366} & 0.8352 & \cellcolor{gray!25}{0.8903} & \cellcolor{gray!25}{0.8970} & \cellcolor{gray!25}{0.8966} & 0.8992 \\
        300 & 0.4421 & 0.4935 & \cellcolor{gray!25}{0.5253} & 0.5287 & 0.8276 & \cellcolor{gray!25}{0.8387} & \cellcolor{gray!25}{0.8449} & 0.8438 & 0.8901 & \cellcolor{gray!25}{0.9001} & \cellcolor{gray!25}{0.9004} & 0.9031 \\
}

%% file: appendix_majority_tables.tex
\begin{table}[H]
\centering
{\fontsize{8}{10}\selectfont
\setlength{\tabcolsep}{5pt}
\resizebox{\textwidth}{!}{%
\begin{tabular}{@{}l*{14}{N}@{}}
\toprule
& \multicolumn{2}{c}{GSM8K}
& \multicolumn{2}{c}{CSQA}
& \multicolumn{2}{c}{HotpotQA}
& \multicolumn{2}{c}{Countdown}
& \multicolumn{2}{c}{GPQA}
& \multicolumn{2}{c}{MBPP}
& \multicolumn{2}{c}{Average} \\
\cmidrule(lr){2-3}
\cmidrule(lr){4-5}
\cmidrule(lr){6-7}
\cmidrule(lr){8-9}
\cmidrule(lr){10-11}
\cmidrule(lr){12-13}
\cmidrule(lr){14-15}
Method
& M@16 & M@32
& M@16 & M@32
& M@16 & M@32
& M@16 & M@32
& M@16 & M@32
& M@16 & M@32
& M@16 & M@32 \\
\midrule
\multicolumn{15}{l}{\textbf{Qwen2.5-1.5B-Instruct}} \\
\global\softtabdatatrue
Base
& 82.6 & 84.3
& 67.8 & 68.8
& \underline{31.1} & \underline{31.1}
& 21.7 & 27.0
& 20.7 & 20.2
& \underline{61.9} & 61.5
& 47.6 & 48.8 \\
\grpomethod
& \underline{84.2} & \underline{84.8}
& \underline{70.0} & \underline{71.0}
& 28.8 & 29.0
& \underline{27.8} & 31.5
& 22.7 & 22.7
& 60.3 & 61.9
& \textbf{49.0} & \cellcolor{gray!35}50.2 \\
\esmethod
& 82.3 & 83.9
& 67.2 & 68.9
& 29.8 & 29.8
& 27.6 & \underline{33.1}
& \underline{25.3} & \underline{24.2}
& 58.4 & \underline{62.3}
& \cellcolor{gray!35}48.4 & \textbf{50.4} \\
\esmethod$\rightarrow$\grpomethod
& 83.9 & 84.2
& 69.2 & 69.6
& 28.8 & 29.1
& 26.2 & 31.6
& 20.7 & 18.7
& 61.5 & 62.7
& 48.4 & 49.3 \\
\grpomethod$\rightarrow$\esmethod
& 83.1 & 83.6
& 68.7 & 70.0
& 29.2 & 29.4
& 20.7 & 24.6
& 23.7 & 23.7
& 58.8 & 61.9
& 47.4 & 48.9 \\
\midrule
\multicolumn{15}{l}{\textbf{Llama-3.2-3B-Instruct}} \\
Base
& 86.9 & 87.8
& 74.1 & 74.6
& 28.1 & 28.5
& 39.2 & 45.7
& 33.8 & \underline{33.8}
& \underline{62.3} & \underline{64.6}
& 54.1 & 55.8 \\
\grpomethod
& \underline{88.9} & \underline{89.5}
& 74.3 & 74.6
& 27.4 & 27.7
& \underline{44.6} & \underline{50.8}
& \underline{35.4} & \underline{33.8}
& \underline{62.3} & 63.0
& \textbf{55.5} & \textbf{56.6} \\
\esmethod
& 88.6 & 88.9
& \underline{74.9} & \underline{75.5}
& \underline{29.5} & \underline{30.0}
& 40.4 & 48.9
& 30.3 & 30.8
& \underline{62.3} & 61.9
& \cellcolor{gray!35}54.4 & \cellcolor{gray!35}56.0 \\
\esmethod$\rightarrow$\grpomethod
& 88.4 & 88.2
& 75.9 & 75.8
& 26.6 & 26.8
& 41.9 & 48.7
& 29.8 & 29.3
& 63.8 & 62.3
& \cellcolor{gray!35}54.4 & 55.2 \\
\grpomethod$\rightarrow$\esmethod
& 89.1 & 89.8
& 74.9 & 75.6
& 28.5 & 28.6
& 40.0 & 47.5
& 30.8 & 31.3
& 63.0 & 62.7
& 54.4 & 55.9 \\
\midrule
\multicolumn{15}{l}{\textbf{Qwen2.5-7B-Instruct}} \\
Base
& \underline{94.5} & \underline{94.5}
& 81.4 & \underline{81.7}
& 45.9 & 46.0
& 56.6 & 58.4
& 37.4 & 38.4
& \underline{77.0} & \underline{77.0}
& 65.5 & \cellcolor{gray!35}66.0 \\
\grpomethod
& 93.8 & 93.6
& \underline{81.9} & 81.4
& \underline{46.8} & \underline{46.9}
& 56.6 & 57.4
& 34.9 & 37.9
& 75.9 & 75.9
& 65.0 & 65.5 \\
\esmethod
& 94.2 & 94.2
& 80.8 & 81.2
& 45.7 & 45.7
& \underline{57.9} & \underline{60.1}
& \underline{40.4} & \underline{41.9}
& 75.1 & 75.9
& \textbf{65.7} & \textbf{66.5} \\
\esmethod$\rightarrow$\grpomethod
& 94.3 & 94.6
& 81.4 & 81.4
& 46.5 & 46.5
& 57.3 & 59.2
& 37.4 & 37.9
& 75.1 & 74.7
& 65.3 & 65.7 \\
\grpomethod$\rightarrow$\esmethod
& 93.7 & 94.0
& 82.2 & \underline{81.7}
& 46.2 & 46.3
& 56.1 & 57.2
& \underline{40.4} & 39.4
& 75.1 & 75.9
& \cellcolor{gray!35}65.6 & 65.8 \\
\bottomrule
\end{tabular}%
}
\global\softtabdatafalse
\par}
\caption{Maj@\(K\) results (\(\times 100\)) in the Easy Setting after two epochs of GSM8K post-training, corresponding to the Pass@\(K\) results in Table~\ref{tab:rq1-instruct-final}. Underlining marks the best Base/GRPO/ES result in each task cell; bold and shading mark the best and second-best Average cells across all five methods.}
\label{tab:app-rq1-instruct-majority}
\end{table}

\begin{table}[H]
\centering
{\fontsize{8}{10}\selectfont
\setlength{\tabcolsep}{12pt}
\resizebox{\textwidth}{!}{%
\begin{tabular}{@{}l*{10}{N}@{}}
\toprule
& \multicolumn{2}{c}{AIME24}
& \multicolumn{2}{c}{AIME25}
& \multicolumn{2}{c}{AMC23}
& \multicolumn{2}{c}{MATH500}
& \multicolumn{2}{c}{Average} \\
\cmidrule(l{1pt}r{1pt}){2-3}
\cmidrule(l{1pt}r{1pt}){4-5}
\cmidrule(l{1pt}r{1pt}){6-7}
\cmidrule(l{1pt}r{1pt}){8-9}
\cmidrule(l{1pt}r{1pt}){10-11}
Method
& M@16 & M@32
& M@16 & M@32
& M@16 & M@32
& M@16 & M@32
& M@16 & M@32 \\
\midrule
\multicolumn{11}{l}{\textbf{DeepSeek-R1-Distill-Qwen-1.5B}} \\
\global\softtabdatatrue
Base
& 36.7 & 40.0
& 33.3 & \underline{33.3}
& 82.5 & 82.5
& 90.0 & 91.0
& 60.6 & 61.7 \\
\grpomethod
& \underline{43.3} & \underline{46.7}
& 30.0 & \underline{33.3}
& 85.0 & 87.5
& 91.2 & 91.8
& 62.4 & 64.8 \\
\esmethod
& 36.7 & 43.3
& \underline{36.7} & \underline{33.3}
& \underline{90.0} & \underline{92.5}
& \underline{91.8} & \underline{92.2}
& 63.8 & \cellcolor{gray!35}65.3 \\
\esmethod$\rightarrow$\grpomethod
& 46.7 & 53.3
& 33.3 & \underline{33.3}
& 85.0 & \underline{92.5}
& 91.6 & 92.6
& \cellcolor{gray!35}64.2 & \textbf{67.9} \\
\grpomethod$\rightarrow$\esmethod
& 46.7 & \underline{46.7}
& 40.0 & 36.7
& \underline{90.0} & 85.0
& 91.4 & \underline{92.2}
& \textbf{67.0} & 65.1 \\
\bottomrule
\end{tabular}%
}
\global\softtabdatafalse
\par}
\caption{Maj@\(K\) results (\(\times 100\)) on mathematical benchmarks in the Hard Setting, corresponding to the Pass@\(K\) results in Table~\ref{tab:rq1-deepseek-math}. Underlining marks the best Base/GRPO/ES result in each task cell; bold and shading mark the best and second-best Average cells across all five methods.}
\label{tab:app-rq1-deepseek-math-majority}
\end{table}

\begin{table}[H]
\centering
{\fontsize{8}{10}\selectfont
\setlength{\tabcolsep}{1.2pt}
\begin{tabular*}{\textwidth}{@{\extracolsep{\fill}}l*{10}{N}@{}}
\toprule
& \multicolumn{2}{c}{GPQA}
& \multicolumn{2}{c}{MBPP}
& \multicolumn{2}{c}{CSQA}
& \multicolumn{2}{c}{Countdown}
& \multicolumn{2}{c}{Average} \\
\cmidrule(l{1pt}r{1pt}){2-3}\cmidrule(l{1pt}r{1pt}){4-5}
\cmidrule(l{1pt}r{1pt}){6-7}\cmidrule(l{1pt}r{1pt}){8-9}
\cmidrule(l{1pt}r{1pt}){10-11}
Method & M@16 & M@32
& M@16 & M@32
& M@16 & M@32
& M@16 & M@32
& M@16 & M@32 \\
\midrule
\multicolumn{11}{l}{\textbf{DeepSeek-R1-Distill-Qwen-1.5B}} \\
\global\softtabdatatrue
Base & 33.3 & 33.3 & 69.7 & 70.4 & \underline{54.1} & \underline{54.1} & 71.1 & 75.1 & 57.0 & 58.2 \\
\grpomethod & 35.9 & 35.4 & 69.7 & \underline{70.8} & 53.7 & 52.7 & 68.5 & 72.8 & 56.9 & 57.9 \\
\esmethod & \underline{39.4} & \underline{39.4} & \underline{70.8} & 69.7 & 52.7 & 53.6 & \underline{71.7} & \underline{76.2} & \textbf{58.6} & \textbf{59.7} \\
\esmethod$\rightarrow$\grpomethod & 36.4 & 37.4 & 71.6 & 72.0 & 52.4 & 53.0 & 70.9 & 73.6 & \cellcolor{gray!35}57.8 & \cellcolor{gray!35}59.0 \\
\grpomethod$\rightarrow$\esmethod & 33.8 & 34.9 & 66.5 & 69.7 & 53.3 & 54.1 & 72.2 & 76.9 & 56.5 & 58.9 \\
\bottomrule
\end{tabular*}
\global\softtabdatafalse
\par}
\caption{Maj@\(K\) results (\(\times 100\)) on four held-out benchmarks in the one-epoch Hard Setting, corresponding to the Pass@\(K\) results in Table~\ref{tab:rq2-deepseek-nonmath}. Underlining marks the best Base/GRPO/ES result in each task cell; bold and shading mark the best and second-best Average cells across all five methods.}
\label{tab:app-rq2-deepseek-nonmath-majority}
\end{table}